\documentclass[letterpaper, 10 pt, conference]{ieeeconf}  

\IEEEoverridecommandlockouts                              

\usepackage{amssymb}
\usepackage{graphicx}
\usepackage{amsmath} 
\usepackage{booktabs}
\usepackage{comment}
\usepackage{hyperref}
\usepackage{xcolor}
\usepackage{algorithm}
\usepackage{algpseudocode}
\usepackage{setspace}  
\usepackage{tabularx}
\usepackage{placeins}

\definecolor{darkgreen}{rgb}{0.0, 0.6, 0.0}
\definecolor{darkred}{rgb}{0.9, 0.35, 0.35} 
\definecolor{mygray}{gray}{0.55}

\makeatletter
\newcommand\fs@betterruled{%
  \def\@fs@cfont{\bfseries}\let\@fs@capt\floatc@ruled
  \def\@fs@pre{\vspace*{7.2pt}\hrule height.8pt depth0pt \kern2pt}%
  \def\@fs@post{\kern2pt\hrule\relax}%
  \def\@fs@mid{\kern2pt\hrule\kern2pt}%
  \let\@fs@iftopcapt\iftrue}
\floatstyle{betterruled}
\restylefloat{algorithm}
\makeatother

\newcommand{\dropbox}{\mbox{box-v1}}
\newcommand{\doll}{\mbox{doll-v1}}
\newcommand{\pick}{\mbox{pick-v1}} 
\newcommand{\push}{\mbox{push-v1}}
\newcommand{\pickv}{\mbox{pick-wrist-v1}} 
\newcommand{\picksim}{\mbox{sim-pick-v1}} 
\newcommand{\pickvv}{\mbox{pick-v2}} 
\newcommand{\pizero}{$\pi_0$}

\title{\LARGE \bf
Augmented Reality for RObots (ARRO): Pointing Visuomotor Policies Towards Visual Robustness
}

\author{
Reihaneh Mirjalili$^{*1}$, Tobias Jülg$^{*1}$, Florian Walter$^{1}$, Wolfram Burgard$^{1}$ \\
\thanks{$^*$Equal contribution $^{1}$All authors are with the Department of Computer Science and Artificial Intelligence, University of
Technology Nuremberg, Nuremberg, Germany.}
}

\begin{document}

\maketitle
\thispagestyle{empty}
\pagestyle{empty}

\begin{abstract}

    Visuomotor policies trained on human expert demonstrations have recently shown strong performance across a wide range of robotic manipulation tasks.
    However, these policies remain highly sensitive to domain shifts stemming from background or robot embodiment changes, which limits their generalization capabilities.
    In this paper, we present ARRO, a novel visual representation that leverages zero-shot open-vocabulary segmentation and object detection models to efficiently mask out task-irrelevant regions of the scene in real time without requiring additional training, modeling of the setup, or camera calibration.
    By filtering visual distractors and overlaying virtual guides during both training and inference, ARRO improves robustness to scene variations and reduces the need for additional data collection.
    We extensively evaluate ARRO with Diffusion Policy on a range of tabletop manipulation tasks in both simulation and real-world environments, and further demonstrate its compatibility and effectiveness with generalist robot policies, such as Octo, \mbox{OpenVLA} and \pizero.
    Across all settings in our evaluation, ARRO yields consistent performance gains, allows for selective masking to choose between different objects, and shows robustness even to challenging segmentation conditions.
    Videos showcasing our results are available at: \href{https://augmented-reality-for-robots.github.io/}{augmented-reality-for-robots.github.io}

\end{abstract}

\section{INTRODUCTION}
\bstctlcite{IEEEexample:BSTcontrol}
\label{sec:introduction}
Visuomotor policy learning in robotics has recently benefited significantly from advances in generative modeling \cite{Vaswani.2017, Ho.2020, Dosovitskiy.2021}. State-of-the-art methods for imitation learning from human expert demonstrations show strong performance in both tabletop and mobile manipulation tasks \cite{visuomotordiffusion, viola, dalal2023imitating}, which has motivated the collection of a vast range of robotics datasets \cite{Walke.2023, Fu.2024, roboagent, Fang.2024, droiddataset, robomind} and concerted efforts to curate and share them~\cite{openx}. This has enabled the development and training of large-scale visuomotor policies that are commonly referred to as robotics foundation models and cover a wide range of tasks, robots, and environments \cite{openx, rt1, rt2, octo, openvla, pizero}.

Ideally, visuomotor policies should be robust to visual environment changes and agnostic to the robot's appearance. However, despite the generalization capabilities that some of these policies achieve in specific scenarios, generalization under domain shift remains a challenge. Even subtle visual changes---such as background variation, distractor objects, or differences in the appearance of the robot---that occur when a policy is deployed in conditions different from those seen during training can lead to notable performance degradation~\cite{decomposegengap, Li.2024}. This severely limits the applicability of such policies in real-world environments, where visual variability is inevitable. Although increasing the diversity of the training data can partially mitigate these issues, such strategies are costly and often fail to cover the full spectrum of real-world variation.

\begin{figure}[t]
    \centering
    \includegraphics[width=0.8\linewidth, trim=0 0 0 0, clip]{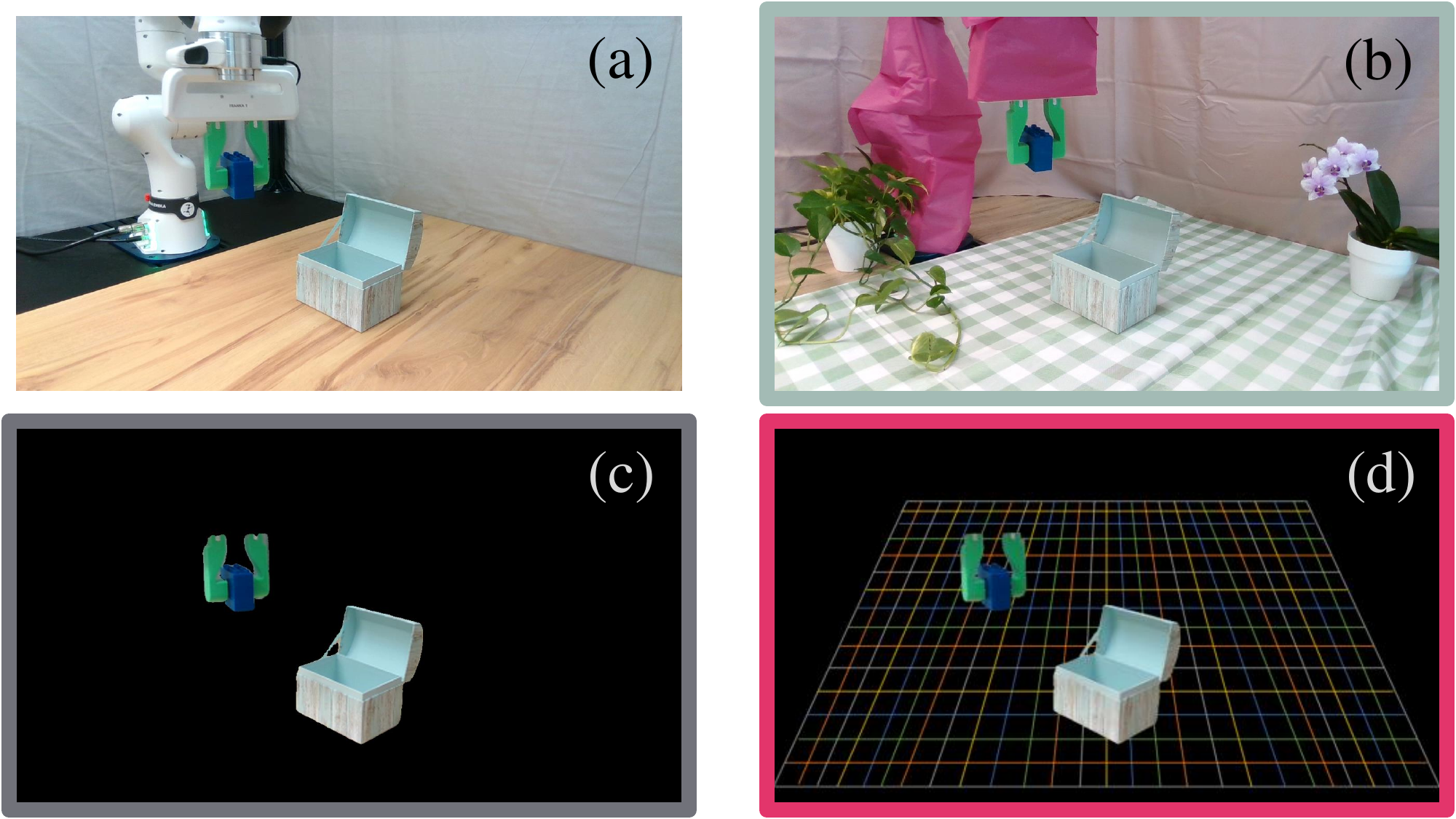}\\
    \vspace*{0.5em}
    \includegraphics[width=0.9\linewidth, trim=0 0 0 0, clip]{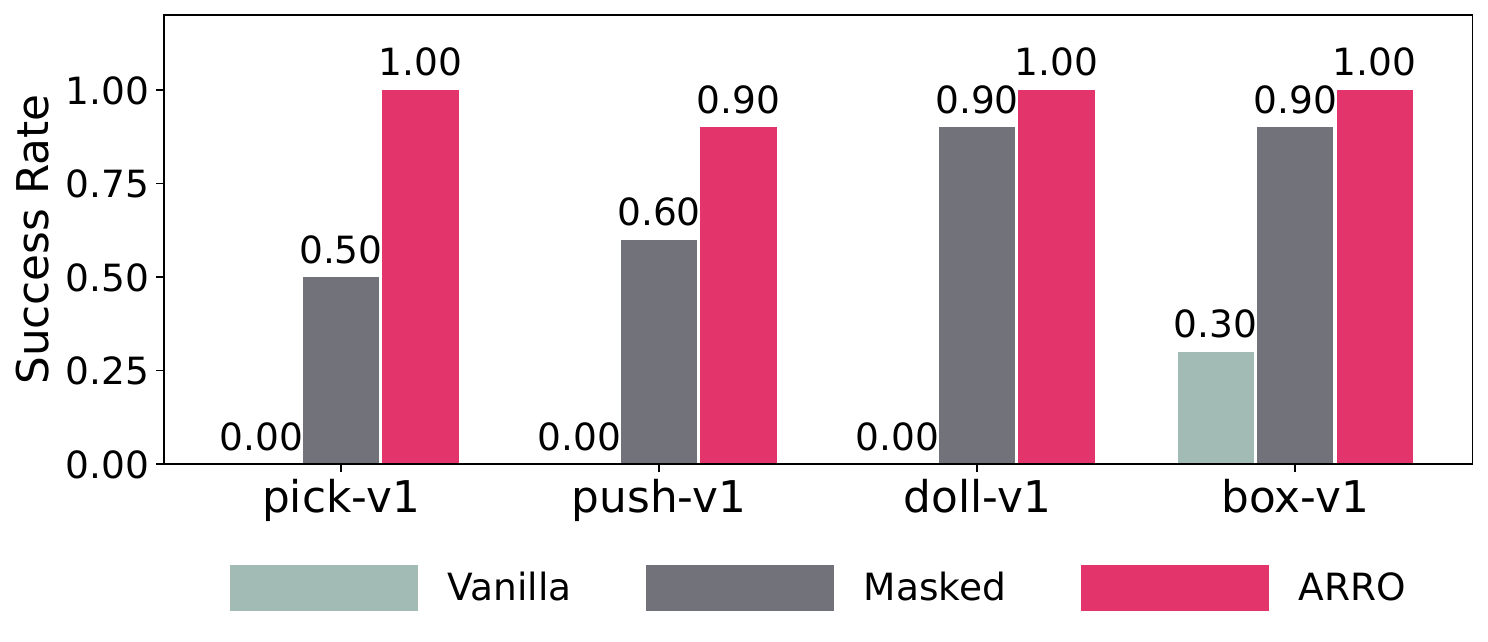}
    \caption{Visualization of input formats (top) and performance comparison across four manipulation tasks (bottom). (a) shows the training scene, while (b) depicts an altered scene with visual domain shifts used during inference. (c) and (d) illustrate the corresponding inputs for the masked diffusion policy and ARRO, respectively.}
    \label{fig:bar_plot}
\end{figure}

To address these challenges, we propose a shift in perspective motivated by recent developments in foundation models for computer vision. Instead of training policies to handle every possible visual scenario, we aim to transform the input space to a canonical representation that cancels out scene variations not relevant for the task. Concretely, we ask: 

\textit{Can we equip robots with a task-oriented augmented reality view that selectively filters irrelevant information and emphasizes only what is essential for task execution?}

In this paper, we introduce \textbf{ARRO} (\textbf{A}ugmented \textbf{R}eality for \textbf{RO}bots), a method that functions like \textit{AR glasses} for robots—minimizing visual distractions, highlighting task-relevant elements, and enhancing policy robustness without requiring retraining. ARRO is a calibration-free visual pre-processing pipeline for generating task-specific augmented visual observations. By leveraging open-vocabulary segmentation and object detection, it retains only the robot gripper and the target objects, and overlays them onto a structured virtual background. This process results in a consistent, simplified input space that supports visuomotor robustness across varied environments and embodiments.

\textbf{In summary, we make the following contributions:} \textbf{(1)} We propose ARRO, a calibration-free augmented reality pipeline that improves the robustness of visuomotor policies by selectively retaining task-relevant visual information---specifically, the fingers of the robot's gripper and manipulated objects---while masking out distractors. \textbf{(2)} We introduce ARRO's system design, which is based on vision-language models and computer vision foundation models for segmentation and object detection. As a result, it operates in an open-vocabulary and zero-shot manner. Neither camera calibration, nor modeling of the task or environment, nor training are required. In principle, ARRO can be combined with any visuomotor policy. \textbf{(3)} We create a virtual background with clear visual references that is included in the virtual scene. Our experimental results show that this improves performance compared to a uniform background without visual markers. \textbf{(4)} We evaluate ARRO in multiple tasks and settings on Diffusion Policy \cite{visuomotordiffusion}, Octo \cite{octo}, OpenVLA \cite{openvla} and \pizero{}~\cite{pizero}. Our results show improved robustness to background variations, distractors, and highlight how ARRO enables cross-embodiment transfer.

\section{Related Works}
\label{sec:related}
\textbf{Domain Adaptation in Imitation Learning}
Domain adaptation for visuomotor robot policies has been explored in different contexts. In reinforcement learning, it is especially relevant when transferring a policy trained in simulation to the real world and can be addressed, for example, through domain randomization \cite{Tobin.2017, chebotar2019closing}. In imitation learning, large-scale datasets have emerged as a key strategy for enabling generalization across varied tasks, embodiments, and environments. The Open X-Embodiment dataset~\cite{openx} consolidates demonstrations from a wide range of robotic platforms and has enabled the training of generalist robot policies~\cite{openx, octo, openvla}. Given the cost of collecting large-scale robot data, several works leverage human videos~\cite{chen2021learning, xiong2021learning, phantom}. Another line of research addresses embodiment differences by using wrist-mounted cameras~\cite{polybot, umi}, which maintain consistent robot-centric viewpoints but cannot benefit from the large amount of available third-person data and are limited in tasks requiring broader scene understanding. In contrast, ARRO operates directly on third-person data without retraining or new data collection.

\textbf{Environment Representations for Visuomotor Manipulation Policies}
Representations of the environment that are robust against domain shift are a long-standing research topic in robot learning. Object-centric representations~\cite{viola, Devin.2018} aim to extract task-relevant objects from the scene while excluding everything else. In contrast, ARRO performs the same operation directly in image space and does not require any training. Transporter networks~\cite{Zeng.2021} extend the idea of object-centric representations by not only identifying relevant objects but also outputting corresponding actions that can be conditioned on language~\cite{Shridhar.2022}. Another line of work leverages keypoints and constraints between them to represent objects and goal poses of manipulation tasks \cite{Sieb.2020, Manuelli.2022}. Recently, foundation models were used to generate both keypoints and constraints automatically~\cite{Huang.2024}. Vision-language models have also been shown to operate entirely without keypoints by reasoning over visual annotations in the camera image~\cite{Nasiriany.2024, fangandliu2024moka}. A disadvantage of keypoint-based methods is that they make implicit assumptions about the types of tasks to be executed. More general approaches learn representations from large datasets, such as human videos~\cite{Nair.2022, Karamcheti.2023} and also support 3D data~\cite{Chen.2024}.

\textbf{Visual Editing}
Most closely related to our approach are methods that apply direct visual editing to RGB camera images. Several works augment training data by synthesizing new backgrounds, tasks, or distractor objects~\cite{roboagent, cacti, rosie, genaug, genima}. \emph{RoboSaGa} leverages saliency extraction to overlay out-of-distribution images adaptively while preserving task-relevant image regions~\cite{saga}. Other approaches focus on adapting visual inputs for cross-embodiment transfer and scene variations. In a method termed \emph{VR-Goggles for Robots}, the authors apply style transfer methods to transform real-world camera images to synthetic images that capture the style of the simulation environment that was used for training~\cite{vr-goggles}. 
\emph{Mirage}~\cite{mirage} enables zero-shot cross-embodiment transfer via cross-painting—masking the target robot, inpainting missing regions, and overlaying rendered source robot images. While effective for robot arm transfer, it requires precise camera calibration, URDF files, and does not handle background shifts. \emph{Shadow}~\cite{shadow} overlays robot segmentation masks for embodiment transfer without new data collection, but still requires policy retraining, static backgrounds, and precise calibration. \emph{RoVi-Aug}~\cite{rovi-aug} augments robot datasets using fine-tuned diffusion models to generate new robot embodiments and camera viewpoints. However, it requires paired data, additional training, and does not address background variation. In contrast, our method filters out irrelevant visual information via open-vocabulary segmentation, achieving robustness to background and embodiment changes without needing calibration, re-training, or extra data.

\section{Approach}
\begin{figure*}[t]
\centering
\vspace{7.2pt}
\includegraphics[clip,trim=0cm 9.2cm 0cm 0cm,width=0.9\linewidth]{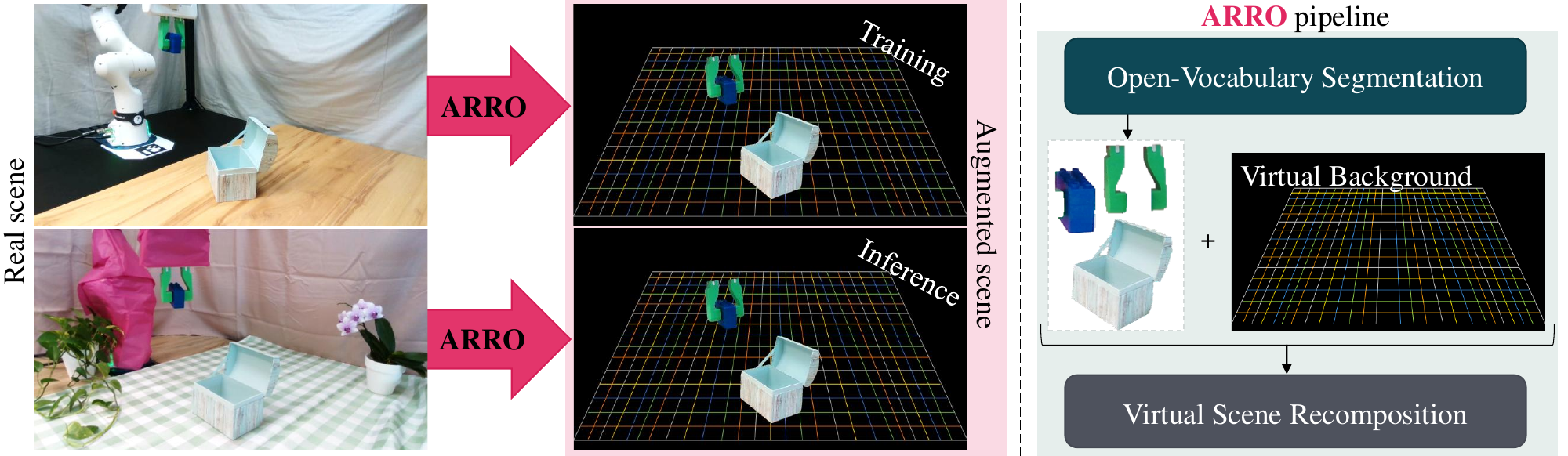}
\caption{ARRO in a nutshell: Our pipeline segments the robot gripper and task-related using open-vocabulary vision models and overlays them onto a virtual background. We consistently use this process across training and inference to enhance robustness against visual domain shifts.}
\label{fig:coverpic}
\end{figure*}

 We adopt the standard visuomotor control setting, where at each time step \( t \), a robot observes a camera frame \( I_t \) and executes an action \( a_t \) based on a policy \linebreak \( \pi(a_t, \ldots, a_{t+T_a} \mid I_{t-T_o}, \ldots, I_t) \) with observation horizon \( T_o \) and action horizon \( T_a \). Throughout this paper, \( I_t \) is an RGB camera image and \( a_t \) an absolute or relative task space action. \( \pi \) is trained on a source-domain dataset of expert demonstrations \( \mathcal{D}_{\text{train}} = \{(I_t, a_t)\} \). When deployed in a target domain \( \mathcal{D}_{\text{test}} \) with novel backgrounds, distractor objects, or changes in robot embodiment, such policies often fail due to sensitivity to task-irrelevant visual variations. Our objective is to compute a calibration-free visual transform \mbox{\( \Phi(I_t) = \tilde{I}_t \)} that takes as input the unaltered image \( I_t \) and outputs the corresponding augmented image \( \tilde{I}_t \), which only contains task-relevant elements.
We design \( \Phi \) based on state-of-the-art vision foundation models to enable zero-shot open-vocabulary object detection without training and apply a calibration-free initialization scheme that leverages a vision-language model. ARRO can thus be deployed without any setup-specific adjustments. Fig. \ref{fig:coverpic} provides an overview of ARRO's processing pipeline.

\subsection{Open-Vocabulary Segmentation}
 To isolate the robot's gripper fingers and task-related objects, we employ a two-phase segmentation process, consisting of an initial segmentation on the first incoming frame, followed by temporally consistent segmentation propagation over time. At the start of an episode, we apply an open-vocabulary object detection model (e.g., Grounding DINO ~\cite{groundingdino}), prompted with object class labels $p^o_1, \dots, p^o_n$, to the first frame \(I_0\), yielding a set of bounding boxes $\mathcal{B}$ with:
\begin{equation}
B_i = \texttt{Detect}(I_0, p^o_i) 
\end{equation}
These bounding boxes can then be used to extract the object with a promptable segmentation model like \mbox{SAM 2}~\cite{sam2}.
Segmenting the gripper fingers requires a different approach. Standard object detection models are not well-suited for detecting robot grippers, as such objects are typically underrepresented in the training data. Therefore, to segment the fingers, we first apply standalone segmentation to \(I_0\), producing a set $\mathcal{K}$ of region proposals without any prompts:
\begin{equation}
    \{K_0, \ldots, K_l\} = \texttt{Segment}(I_0)
\end{equation}
The original image \(I_0\) is then annotated by placing numbered labels at the center of each segmented region. This annotated image, denoted as \( I_0^* \), is passed to a vision-language model (e.g., GPT-4o~\cite{gpt-4o}) along with a task prompt \( p^t \) to identify the regions corresponding to the gripper fingers. This approach also works for objects that have a simple shape and, therefore, can be segmented without a bounding box:
\begin{equation}
\{K^*_0, \ldots, K^*_m\} = \texttt{VLM}(I_0^*, p^t)
\end{equation}
The output $\mathcal{K}^*$ of the model represents the detected keypoints \mbox{$K^*_i = (x_{K^*_i}, y_{K^*_i})$} of the specified objects as well as the left and right gripper fingers. Next, using the object bounding boxes \(\mathcal{B}\) and the keypoints \(\mathcal{K}^*\) as prompts, we apply a memory-based segmentation model to extract the segmented regions from the input image:
\begin{equation}
    S_0^{\mathrm{obj}},\; S_0^{\mathrm{gripper}} = \texttt{Segment}(I_0 \mid \mathcal{B}, \mathcal{K}^*)
\end{equation}
In our experiments, we use SAM 2~\cite{sam2}. Once the initial segmentation has been obtained, we track the object and gripper regions, \(S_t^{\mathrm{obj}}\) and \(S_t^{\mathrm{gripper}}\), across all subsequent frames \(I_t\) for \(t > 0\), by conditioning on the memory accumulated during earlier frames~\cite{sam2}. This enables temporally consistent segmentation over time without requiring re-identification by the vision-language model or additional supervision. The full initialization procedure is summarized in Algorithm \ref{alg:arro_init}.

\subsection{Virtual Scene Recomposition}
Once the relevant segmentations are obtained at each timestep \(t\), we extract the task-relevant regions as described in Algorithm \ref{alg:arro_mask}. After retrieving the segmentation masks for frame $I_t$ at time step $t$ with the initialized model, we compute their union
\(
    S_t = S_t^{\mathrm{obj}} \cup S_t^{\mathrm{gripper}},
\)
where \( S_t^{\mathrm{obj}} \) and \( S_t^{\mathrm{gripper}} \) denote the binary masks for the object and the robot's gripper fingers, respectively.
We then overlay \( S_t \) on a simple black background or a hand-crafted colored grid that is reused across all sequences. While not photorealistic, this background provides visual cues and consistency across frames. The final background-augmented image \(\tilde{I}_t\) is computed as
\begin{equation}
    \tilde{I}_t = S_t \odot I_t + (1 - S_t) \odot I_B,
\end{equation}
where \( I_B \) is the selected background image, and \( \odot \) denotes element-wise multiplication.
This augmentation process is applied to all frames in \( \mathcal{D}_{\text{train}} \) and runs in real time during inference.

\begin{algorithm}[t]
\caption{ARRO Initialization}
\label{alg:arro_init}
\small 
\textbf{Input:} An RGB frame $I$; object prompts $p^o_1, \dots, p^o_n$; task prompt $p^t$; open-vocabulary object detector \texttt{Detect} (e.g., \texttt{GroundingDINO}); uninitialized segmentation model \texttt{Segment} (e.g., \texttt{SAM2}); and a vision-language model \texttt{VLM} (e.g., \texttt{GPT-4o}).

\vspace{0.5em}%
\textbf{Output:} The initialized segmentation model $\texttt{Segment}_{0}$
\vspace{0.5em}%
\begin{algorithmic}
\State \textit{// Get object bounding boxes for complex objects:}
\For{$p^o_i$ in $\{ p^o_1, \ldots, p^o_n \}$}
    \State $B_i \leftarrow \texttt{Detect}(I, p^o_i)$
\EndFor
\vspace{0.5em}%
\State \textit{// Run unprompted segmentation on $I$ to get region masks to\newline // segment simple objects and gripper fingers:}
\vspace{0.25em}%
\State $\{K_0, \ldots, K_l\} \leftarrow \texttt{Segment}(I)$
\vspace{0.5em}%
\State \textit{// Annotate keypoints in $I$ with numeric labels to retrieve $I^*$ and\newline // identify task-relevant keypoints in $I^*$ from $\{K_0, \ldots, K_l\}$:}
\vspace{0.25em}%
\State $\{K^*_0, \ldots, K^*_m\} \leftarrow \texttt{VLM}(I^*, p^t)$
\vspace{0.5em}%
\State \textit{// Initialize and return the segmentation model:}
\State \Return $\texttt{Initialize}(\texttt{Segment}, I, B_0, \ldots, B_n, K^*_0, \ldots, K^*_m)$
\end{algorithmic}
\end{algorithm}

\floatstyle{ruled}
\restylefloat{algorithm}
\begin{algorithm}[t]
\caption{ARRO Masking}
\label{alg:arro_mask}
\small 
\textbf{Input:} RGB frame $I_t$; RGB background image $I_B$; initialized segmentation model $\texttt{Segment}_{t}$ (e.g., \texttt{SAM2}).
\vspace{0.5em}

\textbf{Output:} A masked RGB frame $\tilde{I}_t$ and the updated segmentation model $\texttt{Segment}_{t+1}$
\vspace{0.5em}

\begin{algorithmic}
\State \textit{// Track object and gripper masks:} 
\vspace{0.25em}%
\State $S^{\text{obj}}_t,\; S^{\text{gripper}}_t,\; \texttt{Segment}_{t+1}(I) \leftarrow \texttt{Segment}_{t}(I)$
\vspace{0.5em}%
\State \textit{// Combine masks: }
\vspace{0.25em}%
\State $S_t \leftarrow S^{\text{obj}}_t \cup S^{\text{gripper}}_t$
\vspace{0.5em}%
\State \textit{// Overlay $S$ on virtual background $I_B$ retrieve $\tilde{I}$:}
\vspace{0.25em}%
\State $\tilde{I}_t \leftarrow S_t \odot I_t + (1 - S_t) \odot I_B$
\vspace{0.5em}%
\State \Return $\tilde{I}_t,\; \texttt{Segment}_{t+1}$
\end{algorithmic}
\end{algorithm}

\section{Experimental Results}
\label{sec:experiments}
We evaluate ARRO both in real-world experiments and in simulation with Diffusion Policy \cite{visuomotordiffusion}, Octo~\cite{octo}, OpenVLA~\cite{openvla} and \pizero{}~\cite{pizero}. We collected datasets for training and fine-tuning with manual teleoperation and automated control scripts for which we used the Robot Control Stack~\cite{rcs} on a Franka Research 3 (FR3) robot with custom 3D-printed gripper fingers. Each step of an episode contains the complete robot state, including the end effector pose and a third-person view camera image of the scene, which were used for training. 
For \pizero{} we created a special dataset which also contains wrist-mounted camera images.
All models were trained or fine-tuned using the code released along with their original publications. We selected representative tabletop manipulation tasks to evaluate the performance of our proposed image augmentation pipeline under complex visual conditions. They involve the manipulation of soft deformable objects (e.g., an octopus plush toy), visually rich and colorful textures (e.g., a doll), and dynamic geometric object variations (e.g., closing the lid of a box).

\subsection{Real-World Experiments for Diffusion Policy}
\begin{figure}[t]
\centering
\vspace{7.2pt}
\includegraphics[trim=0cm 0.5cm 0cm 0cm,width=0.8\linewidth]{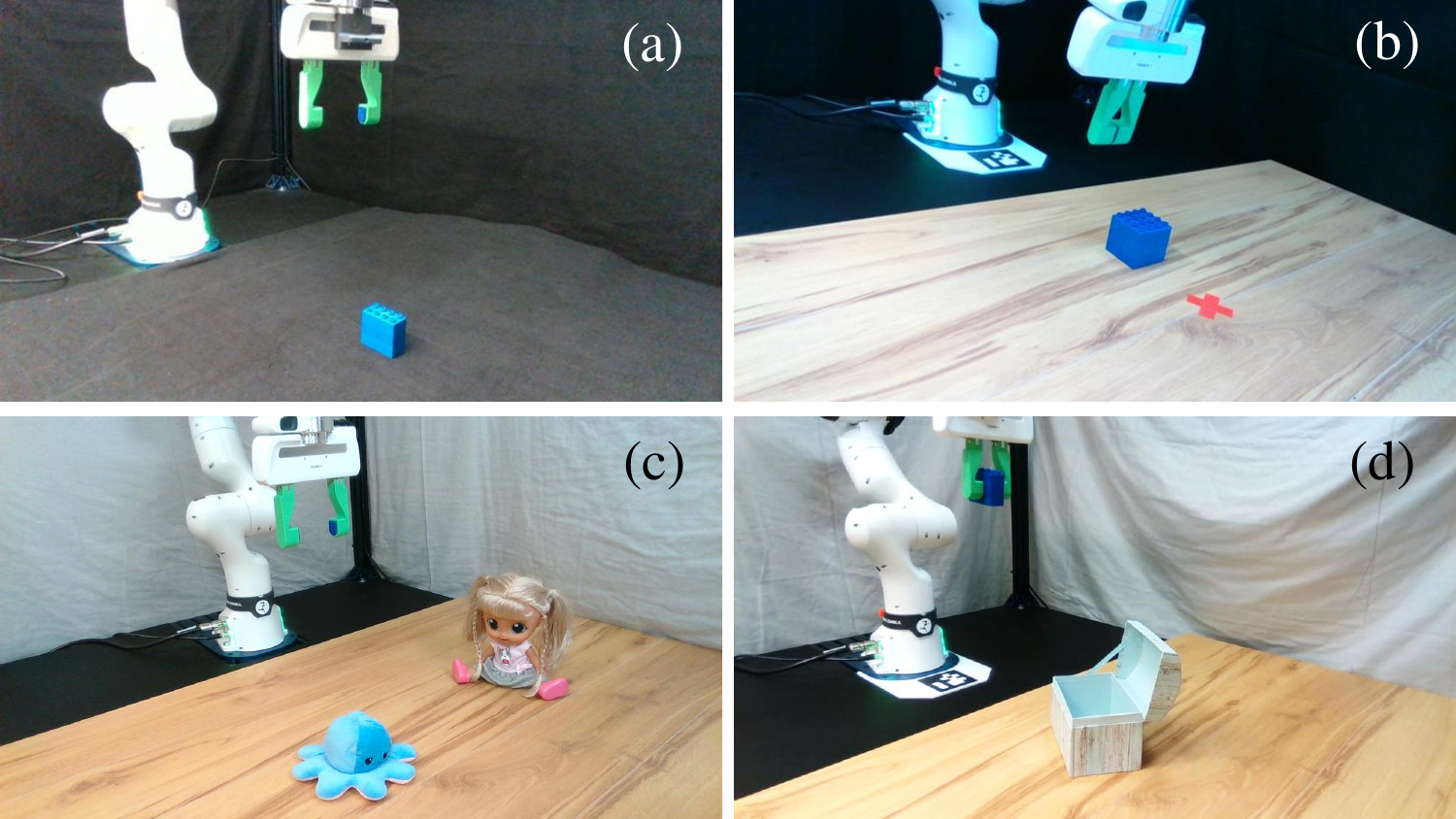}
\caption{Real-world experiment setups for the (a) \pick{}, (b) \push{}, (c)~\doll{} and (d) \dropbox{} tasks.}
\label{fig:thumbnail}
\end{figure}

We conduct four real-world tabletop manipulation tasks that are illustrated in Fig.~\ref{fig:thumbnail}. They include: (a) picking up a single blue cuboid (\pick), (b) pushing a cube to a red cross location (\push), (c) placing a plush octopus next to a doll~(\mbox{\doll}), and (d) dropping a cube into a box and closing the lid (\dropbox{}). The tasks were selected to evaluate a range of manipulation behaviors under realistic visual conditions. For each of them, we collect 90 human demonstrations via teleoperation for training and fine-tuning the policies.

\subsubsection{Performance Across Domain Shifts}
In this section, we evaluate whether ARRO mitigates the degradation in policy performance caused by visual domain shifts. To assess robustness, we introduce significant visual changes at evaluation time, including alterations to the background, modifications to table texture and robot appearance, and the addition of irrelevant objects to the scene. An example of these perturbations for the \dropbox{} task is shown in Fig.~\ref{fig:coverpic}.

\begin{figure}[t]
    \centering
        \centering
        \includegraphics[width=0.8\linewidth, trim=0 250 0 0, clip]
        {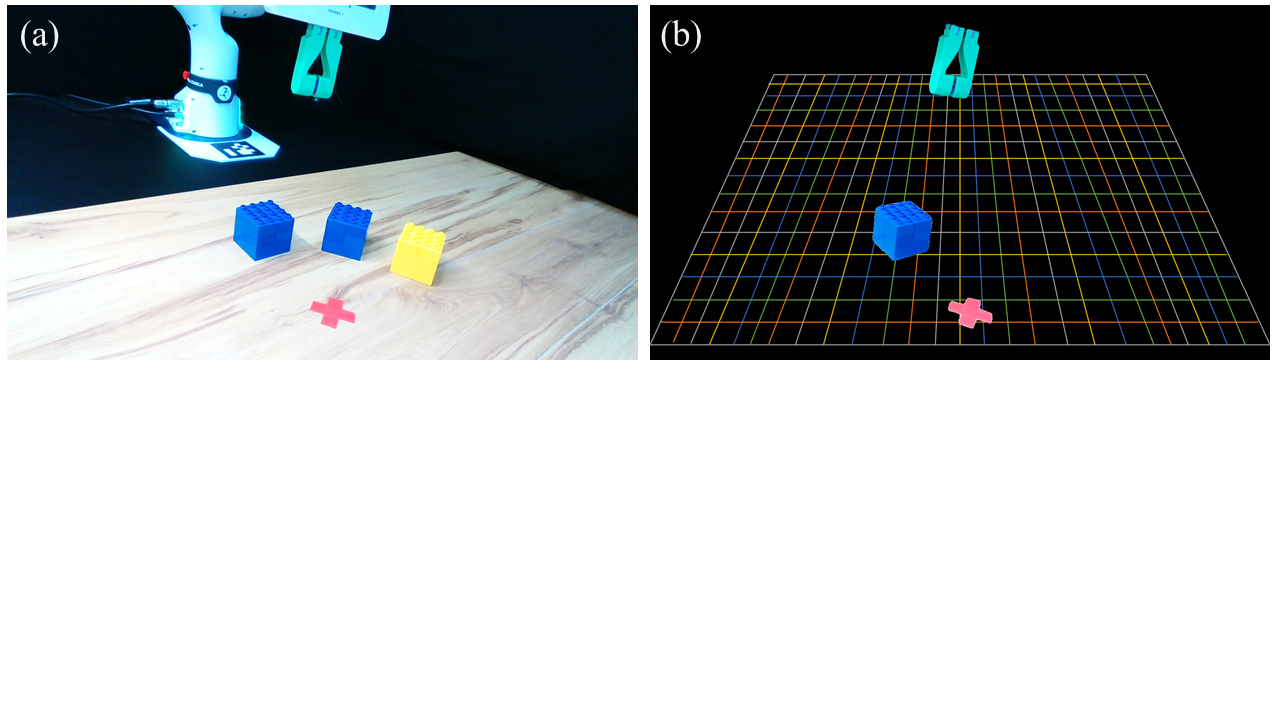}\\[6pt]
        \vfill
         \includegraphics[width=0.75\linewidth, trim=0 25 0 0]
        {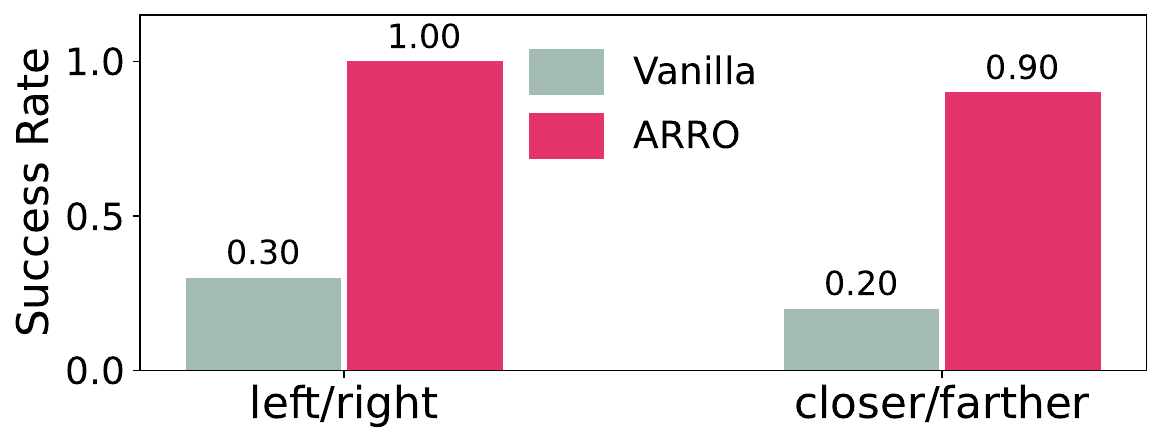}
    \caption{Handling distractor objects. \textit{Top:} example input image with distractors (a) and the ARRO-segmented version (b) where the task is to ``Push the \textit{blue cube that is farther from the yellow cube to} the red cross''. \textit{Bottom:} Success rates for the spatial reasoning tasks.}
    \label{fig:arro-distractor}
\end{figure}

\begin{figure*}[t]
\centering
\vspace{7.2pt}
\includegraphics[trim=0cm 9.6cm 0cm 0cm,width=0.8\linewidth]{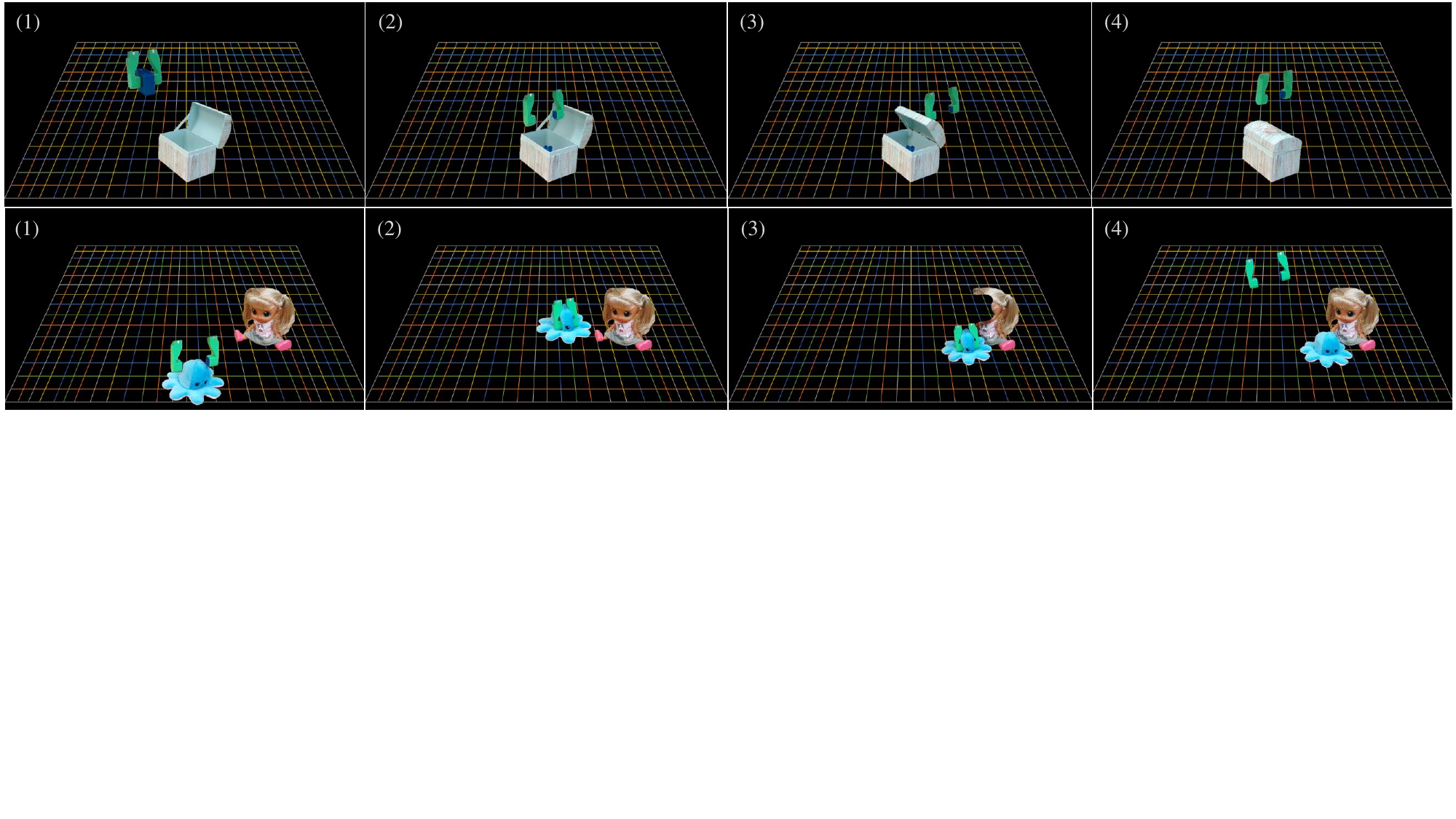}
\caption{Execution sequences using ARRO on the \dropbox{} and \doll{} tasks. By overlaying the segmented task-relevant regions on a virtual background, ARRO neutralizes the effect of visual domain shifts. ARRO's segmentation remains robust to transient occlusions caused by the robot arm or other objects. The doll is temporarily occluded by the octopus plush toy and the gripper in frame (3), but its segmentation accurately reappears once the occlusion clears, without manual correction.}
\label{fig:arro_sequence24}
\end{figure*}

We compare three variants: the \textit{Vanilla Diffusion Policy}, trained and evaluated on unmodified RGB images; the \textit{Masked Diffusion Policy}, which segments the task-relevant objects and places them on a plain black background by masking out all other regions; and \textit{ARRO}, which uses the same segmentation but overlays the task-relevant components onto a structured virtual grid background. Fig.~\ref{fig:bar_plot} illustrates these representations for the \dropbox{} task. For each task and each model, we train the policy for 1000 epochs and evaluate on ten trials.

As shown in Fig.~\ref{fig:bar_plot}, ARRO consistently outperforms both baselines. The Vanilla Diffusion Policy suffers from substantial performance degradation across all scenarios due to its reliance on raw visual features, which no longer remain reliable under domain shifts. The Masked Diffusion Policy partially retains performance by removing distractors. However, it still performs noticeably worse than ARRO. This suggests that while masking suppresses irrelevant information, it may also discard useful visual cues that contribute to task execution. In contrast, ARRO not only filters out distractors but also reintroduces a structured and consistent background, improving robustness and the ability to generalize across varied visual conditions.

To complement the quantitative results, Fig.~\ref{fig:arro_sequence24} presents representative execution sequences using ARRO for the \dropbox{} and \doll{} tasks. Notably, the segmentation module maintains reliable performance even under challenging conditions, such as partial occlusion (e.g., when the doll is partially blocked by the plush toy) or shape deformation (e.g., when the plush toy is grasped or the box begins to close). Despite the presence of visually complex and colorful objects, such as the multi-textured doll, the segmentation remains accurate throughout execution.

\subsubsection{Handling Distractor Objects}
To assess whether ARRO can handle distractor objects and enable spatial reasoning, we repeat the push experiment with additional distractor items in the scene. Specifically, we investigate whether the vision-language model can move beyond basic object recognition (e.g., selecting ``the blue cube'') to spatial grounding (e.g., selecting ``the blue cube farther from the yellow cube'').

We design two spatial reasoning tasks to evaluate whether the system can correctly identify and act upon objects based on spatial relationships. In the first task, the robot must ``push the \textit{blue cube on the left/right} to the red cross.'' In the second, it must ``push the \textit{blue cube that is closer to/farther from the yellow cube} to the red cross.'' Once the vision-language model identifies the correct task-relevant object, ARRO masks out all other image regions, including distractor objects, ensuring that the policy operates solely on the relevant visual input. The resulting task seen by the policy is therefore identical to the \push{} task, which means that no additional data collection or training is required.

As shown in Fig.~\ref{fig:arro-distractor}, ARRO demonstrates strong performance in both tasks and reliably selects the appropriate object based on the spatial relation described in the instruction. In contrast, the vanilla diffusion policy exhibits a substantial drop in performance. It frequently fails to identify the correct object, sometimes selecting a distractor, switching targets mid-execution, or moving ambiguously between multiple candidates. Even when the correct object is selected, the resulting motion is often imprecise. These results highlight ARRO's robustness to distractors and its capacity to support spatially grounded reasoning in visually complex environments.

\subsubsection{Handling Occlusions in ARRO}
Partial occlusions of objects may arise during certain tasks, either due to the robot's own movements or the presence of other objects within the scene, as illustrated in Fig.~\ref{fig:arro_sequence24}. In frame (3), the doll is partially occluded—either by the robot arm or by the octopus plush toy temporarily blocking the view. Despite these transient occlusions, ARRO's segmentation remains stable and accurate. As shown in the figure, once the occlusion subsides, the segmentation of the doll reliably reappears without requiring manual intervention or additional adjustment. This demonstrates that our segmentation pipeline is robust to temporary occlusions and consistently preserves the identification of task-relevant objects over time.

\subsection{Real-World Experiments for Generalist Policies} 
Because ARRO operates directly on raw camera images, it is compatible with any type of visuomotor policy. We therefore explore whether it can also enhance the performance of the language-conditioned generalist policies Octo~\cite{octo}, OpenVLA~\cite{openvla}, and \pizero~\cite{pizero}. While the former two are fine-tuned on \pick{}, the latter is fine-tuned on \pickv{}, which contains episodes for the same task with a green cuboid, but also includes images from a wrist camera mounted close to the robot's gripper. We found that the performance of \pizero{} decreases substantially when it is used with a fixed side camera only.

Using a wrist camera requires additional steps to apply our method, as the background changes when the robot moves, and task-relevant objects may not always be in the camera's field of view. To address this, we always use a black background and apply the ARRO pattern only to the side camera. Moreover, we initialize the segmentation with a pre-recorded image of the scene that shows all relevant objects. We found this method to be robust even if the scene is perturbed. As the gripper fingers always appear at the same initial position in the wrist camera view, we select them directly instead of using the VLM. The wrist camera view can be seen in Fig.~\ref{fig:clutter}.

\subsubsection{Performance Across Domain Shifts}
\begin{figure}[t]
\centering
\vspace{7.2pt}
\includegraphics[trim=0cm 0.9cm 0cm 0cm,width=0.8\linewidth]{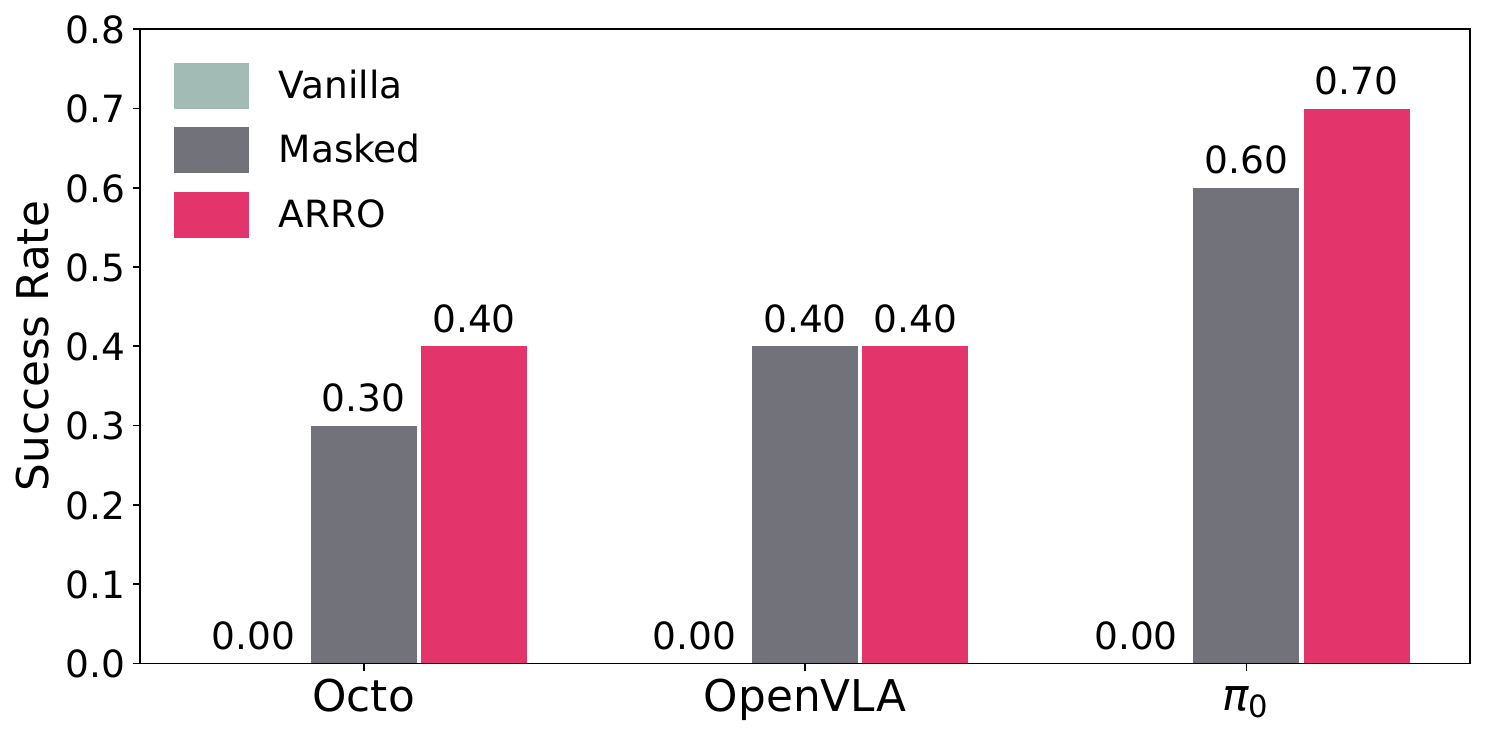}
\caption{Performance comparison across three generalist models, Octo and OpenVLA and \pizero{}, evaluated under altered visual conditions.}
\label{fig:bar_plot_models}
\end{figure}

\begin{figure}[t]
\centering
\vspace{7.2pt}
\includegraphics[width=0.4\linewidth, trim=0 0 0 0cm, clip]{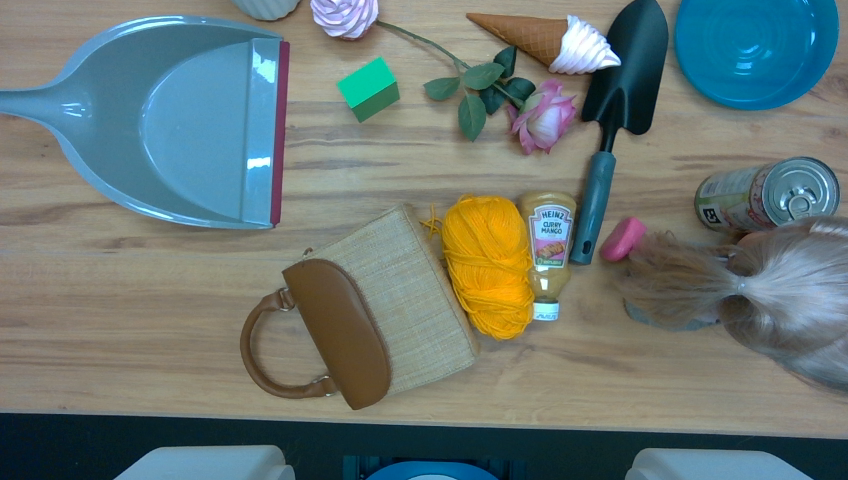}
\hspace{0.2cm}
\includegraphics[width=0.4\linewidth, trim=0 0 0 0, clip]{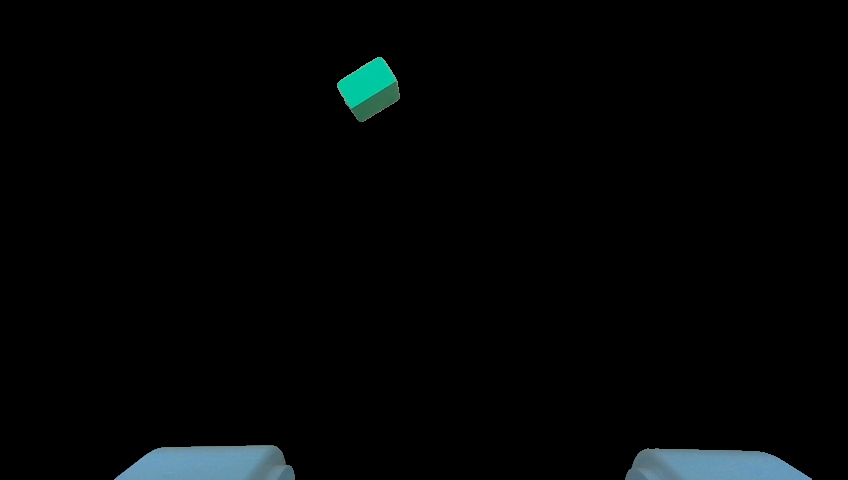}\\[6pt]
\vfill
\includegraphics[width=0.4\linewidth, trim=0 0 0 0, clip]{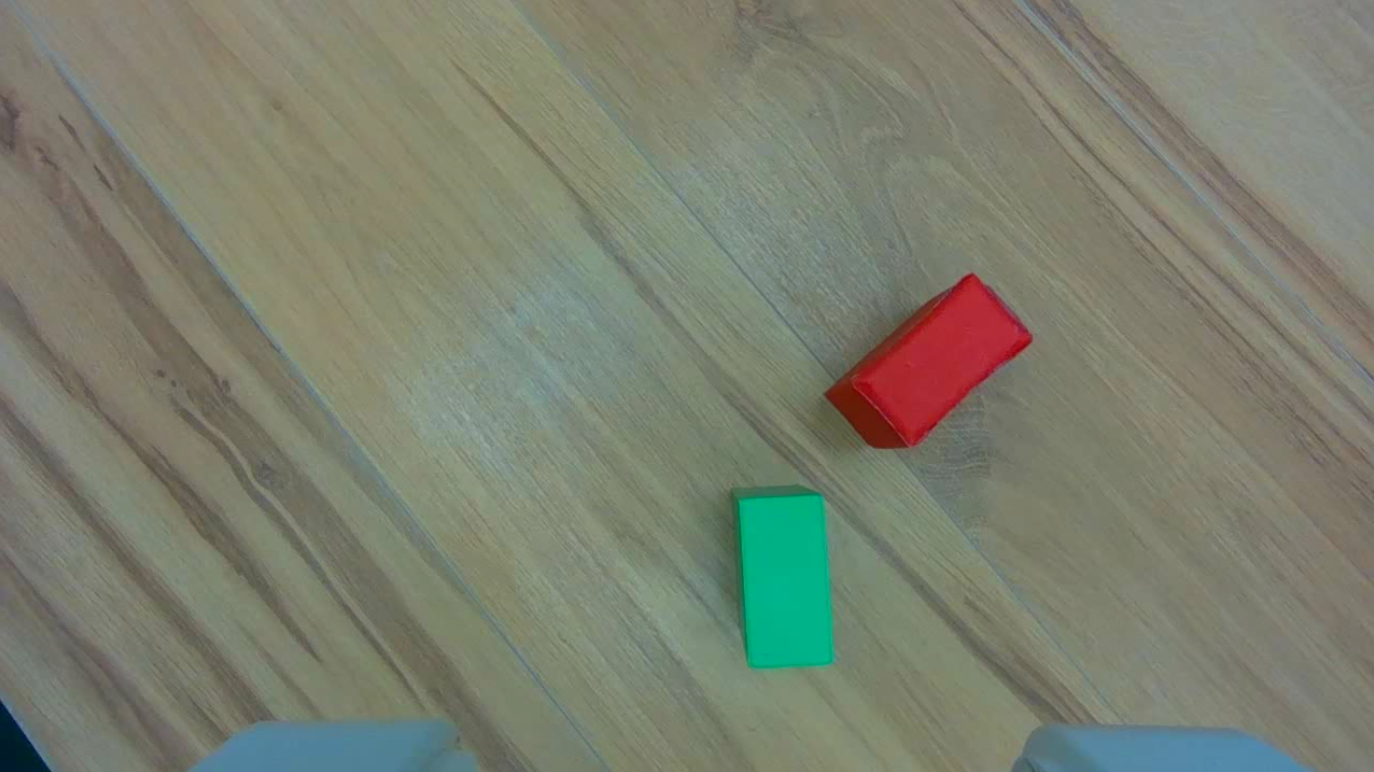}
\hspace{0.2cm}
\includegraphics[width=0.4\linewidth, trim=0 0 0 0, clip]{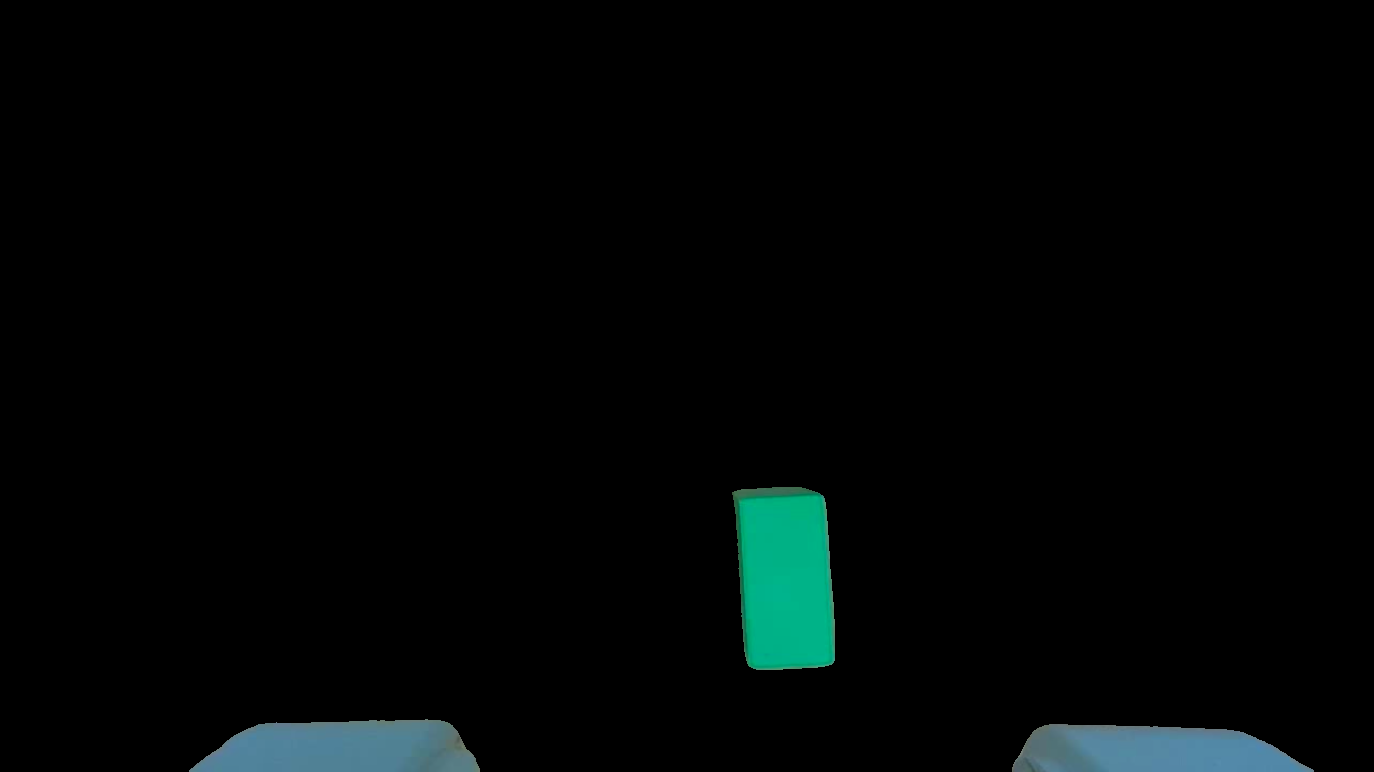}\\
\caption{Wrist camera view of ARRO for the \pizero{} experiment ``pick up the green cube''. \emph{Top}: Example input image of a cluttered scene and the masked version. \emph{Bottom}: Scene with a red cube that is masked by ARRO. }
\label{fig:clutter}
\end{figure}

We evaluated all three models on the pick task under different conditions. The results are presented in Fig.~\ref{fig:bar_plot_models}. As shown in the figure, both the masked and ARRO variants consistently outperform their vanilla counterparts in settings where the visual domain is altered at evaluation time. For Octo and OpenVLA, the degradation in performance of the vanilla variant under domain shifts in some cases results in abrupt and unpredictable behavior. We found \pizero{} to be more robust against visual changes, which is why we added more clutter to the scene, as depicted in Fig.~\ref{fig:clutter}. As shown in Fig.~\ref{fig:bar_plot_models}, this causes the vanilla success rate to drop to 0\%, which can be recovered with both the black masked background and ARRO.

\subsubsection{Language Guidance}
While \pizero{} is more robust against scene changes compared to Octo and OpenVLA, we found that adding an additional red cuboid with the same shape can degrade its performance, as the model ignores the cuboid color in its task instruction.
As a result, the vanilla policy often grasps the wrong cuboid leading to a success rate of 30\%. The masked and ARRO variants, which exclude the task-irrelevant red cube from the scene, increase the success rates to 70\% and 90\%, respectively.

\subsection{ARRO in Simulation}
\begin{figure}[t]
  \centering

  \includegraphics[width=0.3\linewidth, trim=0 90 0 0, clip]{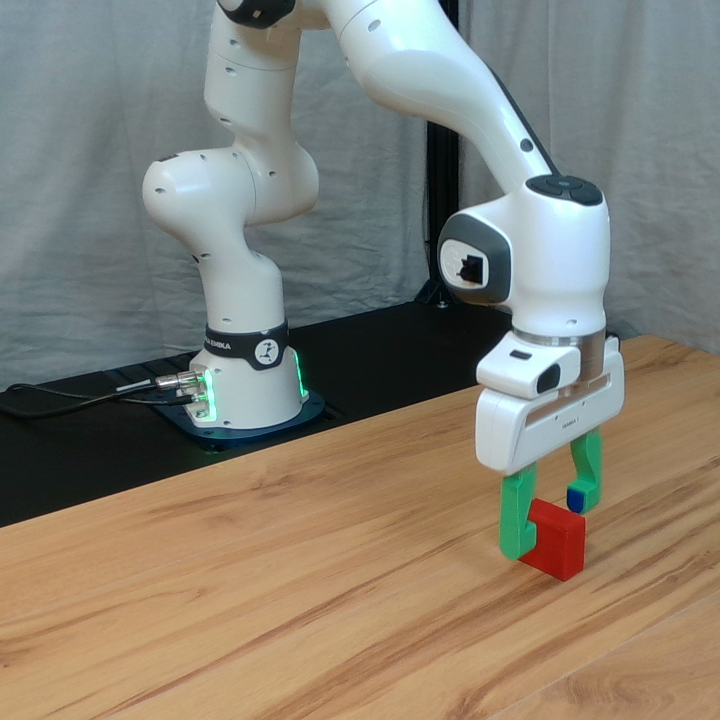}
    \hfill
  \includegraphics[width=0.3\linewidth, trim=0 90 0 0, clip]{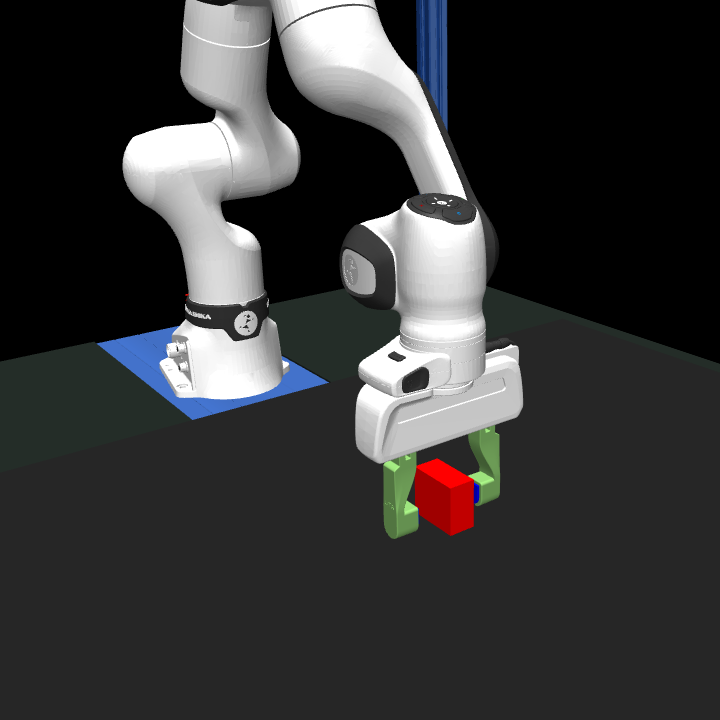}
      \hfill
  \includegraphics[width=0.3\linewidth, trim=0 90 0 0, clip]{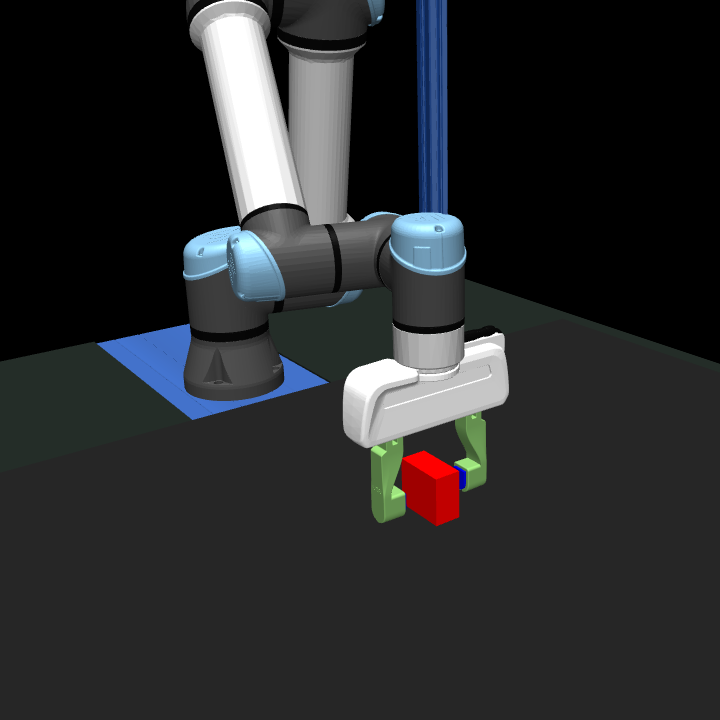}
  \caption{
    Experimental setup for the pick task in two environments and one cross-embodiment scenario. \textit{Left to right}: The real-world FR3 setup, the model of the setup in MuJoCo used for collecting \picksim{}, and the UR5e setup for cross-embodiment experiments in MuJoCo.
  }
  \label{fig:sim_exp_setup}
\end{figure}

In addition to the real-world experiments, we also evaluate ARRO in simulation on the tasks \pickvv{} and \picksim{}. Unlike in \pick{}, the cuboid is red, and we collected the demonstrations automatically via a script in both the real world and the simulation. For \pizero{} we choose \pickv{}, which includes the wrist camera. We investigate whether ARRO can help mitigate the real-to-sim gap~\cite{Li.2024} and facilitate cross-embodiment policy transfer. Fig.~\ref{fig:sim_exp_setup} illustrates our experimental setup: the physical setup in our lab using the FR3 robot, its replication in a \mbox{MuJoCo}~\cite{mujoco} simulation environment, and a cross-embodiment variant where the FR3 is replaced by a UR5e robot in the same environment.

\subsubsection{Real-to-Sim}
\label{sec:real-to-sim}
\begin{figure*}[t]
  \centering
  \includegraphics[trim={0 0.8cm 0 0}, width=0.9\linewidth]{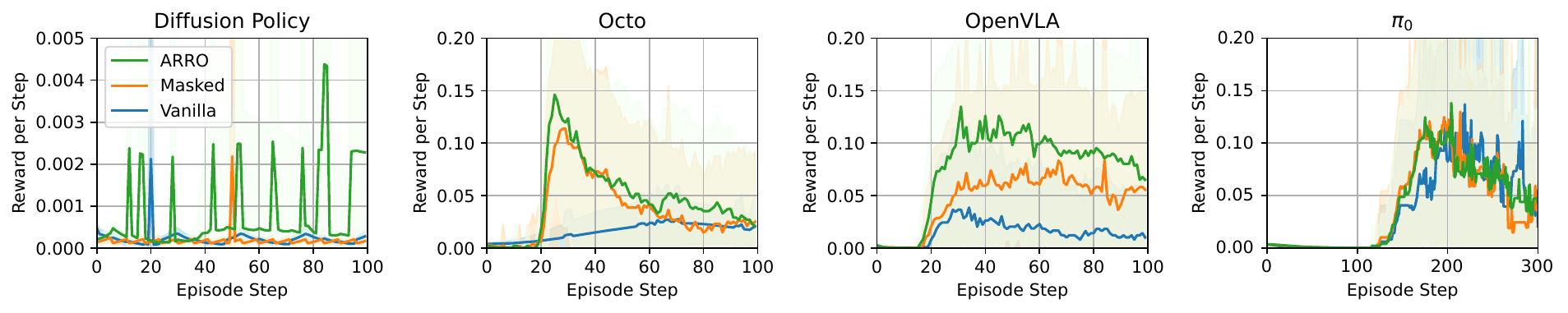}
  \caption{
    Normalized mean per step reward over 100 episodes on the \picksim{} task in simulation on the FR3 (same embodiment).
    All models were trained only on real-world datasets.
    Shaded areas indicate standard deviations.
  }
  \label{fig:real2sim_rewards}
\end{figure*}

Since most datasets for robot imitation learning are collected on real-world setups, we examine a real-to-sim rather than the more typical sim-to-real paradigm to assess ARRO's performance in domain transfer.
We trained Diffusion Policy, Octo and OpenVLA on the \pickvv{} dataset, and \pizero{} on the \pickv{} dataset, and evaluated the task success rates for two settings:
The unmodified real-world scene and the replicated scene in the simulation.
We evaluated the trained policies in-distribution on the same unchanged real-world setup on 10 episodes (real-to-real success rate) and out-of-distribution on the replicated MuJoCo setup on 100 episodes (real-to-sim success rate).
All models achieve decent real-to-real success probabilities in the vanilla case: 90\%, 50\%, 40\%, and 100\% for Diffusion Policy, Octo, OpenVLA, and \pizero{}, respectively.
But as shown in Table~\ref{tab:sim}, the performance drops to 0\% of their real-world performance when evaluated in the real-to-sim setting for all models except for \pizero{}.
With ARRO, OpenVLA retains 55\% of its real-world performance.
Octo does not benefit as much and only achieves 5\% of its original success rate.
Diffusion Policy does not exhibit any immediate transfer in terms of success rate, which can be attributed, at least in part, to the absence of a pre-training stage and differences in the vision backbone architecture.
\pizero{} already has a high baseline performance and can retain 94\% of its success rate when masking is applied.
The low relative performance of ARRO on \pizero{} can be explained by a high real-to-real success probability of 90\%, while the real-to-sim success rate is comparable to that of the masked variant.
In addition to the success rates in the real-to-sim experiments, we computed reward values using a simple distance metric inspired by \mbox{ManiSkill}~\cite{maniskill}. The results are shown in Fig.~\ref{fig:real2sim_rewards}.
For Diffusion Policy, Octo, and OpenVLA, both ARRO and the simple black background yield higher rewards than the vanilla baselines.

\FloatBarrier

\begin{table*}
\centering
\vspace{7.2pt}
\footnotesize
\caption{
Relative policy performance of real-to-sim and cross-embodiment experiments after domain transfer.
}
\label{tab:sim}
\newcolumntype{Y}{>{\centering\arraybackslash}X}
\begin{tabularx}{\linewidth}{lYYlrrrlrrr}
\toprule
 & & & \multicolumn{4}{c}{\textbf{Real-to-Sim}} &  \multicolumn{4}{c}{\textbf{Cross-Embodiment}} \\
\cmidrule(lr){4-7} \cmidrule(lr){8-11}
 \textbf{Policy} & \textbf{Camera} & \textbf{Control}  & \textbf{Dataset} & \textbf{Vanilla} & \textbf{Masked} & \textbf{ARRO} & \textbf{Dataset} & \textbf{Vanilla} & \textbf{Masked} & \textbf{ARRO} \\
\midrule
Diffusion Policy    & Side & Cartesian & \pickvv & 0\% & 0\% &  0\%  & \picksim & 0\% & 80\% & \textbf{99}\% \\
Octo     & Side & Cartesian & \pickvv & 0\% & 3\%  &  \textbf{5}\% & \picksim & 71\% & \textbf{87}\% & \textbf{87}\% \\
OpenVLA  & Side & Cartesian & \pickvv & 0\% & 53\%  &  \textbf{55}\% & \picksim & 29\% & 72\% & \textbf{80}\% \\
\pizero  & Side + Wrist & Joints & \pickv & 79\% & \textbf{94}\% & 77\% & \multicolumn{1}{c}{\text{---}} & \multicolumn{1}{c}{\text{---}} & \multicolumn{1}{c}{\text{---}} & \multicolumn{1}{c}{\text{---}} \\
\bottomrule
\end{tabularx}

\end{table*}

\subsubsection{Cross-Embodiment}
\label{sec:x-embodiment}

To test the cross-embodiment capabilities of ARRO, we train our policies on a simulation dataset, referred to as \picksim{}, using the FR3 and test it on a UR5e embodiment.
We evaluate both in-distribution on the training environment for 100 episodes and on the new UR5e embodiment for 100 episodes.
The relative performance is shown in Table~\ref{tab:sim} under cross-embodiment.
Note that \pizero{} is not included, as it operates in joint space and therefore cannot be directly applied to the UR5e robot due to its different kinematics.

The diffusion policy's performance drops to 0\% when evaluated on a novel embodiment, which is expected given that it was trained exclusively on the FR3 robot.
In contrast, both the masking with a black background and ARRO exhibit only a minor reduction in success rates, retaining most of their performance.
This is likely because the embodiment change has minimal impact on the diffusion policy: the visual input remains largely unaffected due to the masking of the robot body, and the control commands are issued in absolute Cartesian space, which remains invariant to the embodiment.

Octo and OpenVLA also exhibit reduced drops in success rates when combined with ARRO and masking. Unlike the Diffusion Policy, their performance does not drop to 0\% in the vanilla setting for the new embodiment. This can be attributed to their larger and more diverse pre-training datasets, which are explicitly designed to support cross-embodiment generalization. As a result, both models encountered the UR5e robot during training, facilitating better transfer even in simulation settings.

\section{Limitations}
While our method shows clear advantages across all evaluated models and tasks, its performance depends heavily on the underlying segmentation model, especially for wrist cameras, where objects can appear or disappear abruptly. In our experiments, task-relevant objects were occasionally not tracked reliably, leading to degraded performance. This issue could be mitigated by periodically re-initializing the segmentation model or by running a lightweight object detector in parallel to trigger tracker resets for the segmentation model as objects enter or leave the scene. Notably, our approach can still function under minor segmentation errors, depending on the visuomotor policy's tolerance to such perturbations. 

Finally, our method cannot compensate for reflections and changes in lighting. These types of domain shifts could be mitigated with image inpainting methods or integrating depth images. However, the real-to-sim results indicate that more recent models, such as \pizero{}, are less sensitive to such changes.

\section{Conclusion}
\label{sec:conclusion}

We introduce ARRO, a calibration-free visual augmentation pipeline that enhances the robustness of visuomotor policies by segmenting task-relevant regions and compositing them onto a structured virtual background. Our method improves generalization without requiring retraining or camera calibration, and experiments in both real and simulated settings show consistent gains for task-specific as well as generalist policies.

A possible direction for future work is to explore whether, for the wrist-mounted camera, replacing the static background overlay with  dynamic, view-dependent frames would further enhance the results. This remains an open question and a promising area for future investigation.

\section*{ACKNOWLEDGMENT}
The authors acknowledge the HPC resources provided by the Erlangen National HPC Center (NHR@FAU) under the BayernKI project no.~v106be.

\bibliographystyle{IEEEtran}
\bibliography{references}

@string{neurips = "Advances in Neural Information Processing Systems"}

@string{ieeeral = "IEEE Robotics and Automation Letters"}

@string{IROS = "Proc.~of the IEEE/RSJ Int.~Conf.~on Intelligent Robots and Systems (IROS)"}

@string{ICRA = "Proc.~of the IEEE Int.~Conf.~on Robotics \& Automation (ICRA)"}

@string{ECCV = "Proc.~of Europ.~Conf.~on Computer Vision (ECCV)"}

@string{IJRR = "International Journal of Robotics Research"}

@string{ICML = "Proc.~of the Int.~Conf.~on Machine Learning (ICML)"}

@string{IJRR = "Int.~Journal of Robotics Research (IJRR)"}

@string{isrr = "Proc.~of the Int.~Symp.~of Robotics Research (ISRR)"}

@string{rss = "Proc.~of Robotics: Science and Systems (RSS)"}

@string{ICLR = "Proc.~of the Int.~Conf.~on Learning Representations (ICLR)"}

@string{CORL = "Proc.~of the Conf.~on Robot Learning (CoRL)"}

@string{ICCV = "Proc.~of the Int.~Conf.~on Computer Vision (ICCV)"}

@IEEEtranBSTCTL{IEEEexample:BSTcontrol,
CTLuse_forced_etal = "yes",
CTLmax_names_forced_etal = "6",
CTLnames_show_etal = "3",
CTLdash_repeated_names = "no"
}

@InProceedings{shadow,
  title = 	 {SHADOW: Leveraging Segmentation Masks for Cross-Embodiment Policy Transfer},
  author =       {Lepert, Marion and Doshi, Ria and Bohg, Jeannette},
  booktitle = 	 CORL,
  year = 	 {2025},
}

@InProceedings{rovi-aug,
  title = 	 {RoVi-Aug: Robot and Viewpoint Augmentation for Cross-Embodiment Robot Learning},
  author =       {Chen, Lawrence Yunliang and Xu, Chenfeng and Dharmarajan, Karthik and Cheng, Richard and Keutzer, Kurt and Tomizuka, Masayoshi and Vuong, Quan and Goldberg, Ken},
  booktitle = 	 CORL,
  year = 	 {2025}
}

@inproceedings{mirage,
        title={Mirage: Cross-Embodiment Zero-Shot Policy Transfer with Cross-Painting},
        author = {Lawrence Yunliang Chen and Kush Hari and Karthik Dharmarajan and Chenfeng Xu and Quan Vuong and Ken Goldberg},
        booktitle = RSS,
        year = {2024}
}

@inproceedings{cacti,
title={{CACTI}: A Framework for Scalable Multi-Task Multi-Scene  Visual Imitation Learning},
author={Zhao Mandi and Homanga Bharadhwaj and Vincent Moens and Shuran Song and Aravind Rajeswaran and Vikash Kumar},
booktitle={CoRL 2022 Workshop on Pre-training Robot Learning},
year={2022}
}

@misc{rosie,
  title={Scaling robot learning with semantically imagined experience}, 
  author={Yu, Tianhe and Xiao, Ted and Stone, Austin and Tompson, Jonathan and Brohan, Anthony and Wang, Su and Singh, Jaspiar and Tan, Clayton and Peralta, Jodilyn and Ichter, Brian and others},
  year={2023},
  howpublished = {\url{https://arxiv.org/abs/2302.11550}}
}

@misc{genaug,
  title={Genaug: Retargeting behaviors to unseen situations via generative augmentation},
  author={Chen, Zoey and Kiami, Sho and Gupta, Abhishek and Kumar, Vikash},
  howpublished = {\url{https://arxiv.org/abs/2302.06671}},
  year={2023}
}

@InProceedings{genima,
  title = 	 {Generative Image as Action Models},
  author =       {Shridhar, Mohit and Lo, Yat Long and James, Stephen},
  booktitle = 	 CORL,
  year = 	 {2025}
}

@inproceedings{chebotar2019closing,
  title={Closing the sim-to-real loop: Adapting simulation randomization with real world experience},
  author={Chebotar, Yevgen and Handa, Ankur and Makoviychuk, Viktor and Macklin, Miles and Issac, Jan and Ratliff, Nathan and Fox, Dieter},
  booktitle=ICRA,
  year={2019}
}

@article{visuomotordiffusion,
  title={Diffusion policy: Visuomotor policy learning via action diffusion},
  author={Chi, Cheng and Xu, Zhenjia and Feng, Siyuan and Cousineau, Eric and Du, Yilun and Burchfiel, Benjamin and Tedrake, Russ and Song, Shuran},
  journal=ijrr,
  year={2023}
}

@INPROCEEDINGS{chen2021learning, 
    title     = {Learning Generalizable Robotic Reward Functions from ``In-The-Wild'' Human Videos}, 
    author    = {Annie S. Chen AND Suraj Nair AND Chelsea Finn}, 
    booktitle = RSS, 
    year      = {2021}
}

@inproceedings{xiong2021learning,
  title={Learning by watching: Physical imitation of manipulation skills from human videos},
  author={Xiong, Haoyu and Li, Quanzhou and Chen, Yun-Chun and Bharadhwaj, Homanga and Sinha, Samarth and Garg, Animesh},
  booktitle=IROS,
  year={2021},
}

@InProceedings{polybot,
  title = 	 {Polybot: Training One Policy Across Robots While Embracing Variability},
  author =       {Yang, Jonathan Heewon and Sadigh, Dorsa and Finn, Chelsea},
  booktitle = 	 CORL,
  year = 	 {2023}
}

@misc{phantom,
  title={Phantom: Training Robots Without Robots Using Only Human Videos},
  author={Lepert, Marion and Fang, Jiaying and Bohg, Jeannette},
  howpublished = {\url{https://arxiv.org/abs/2503.00779}},
  year={2025}
}

@INPROCEEDINGS{umi,
    title     = {Universal Manipulation Interface: In-The-Wild Robot Teaching Without In-The-Wild Robots}, 
    author    = {Cheng Chi AND Zhenjia Xu AND Chuer Pan AND Eric Cousineau AND Benjamin Burchfiel AND Siyuan Feng AND Russ Tedrake AND Shuran Song},
    booktitle = RSS, 
    year      = {2024}
}

@inproceedings{openx,
  title={Open x-embodiment: Robotic learning datasets and rt-x models: Open x-embodiment collaboration 0},
  author={O’Neill, Abby and Rehman, Abdul and Maddukuri, Abhiram and Gupta, Abhishek and Padalkar, Abhishek and Lee, Abraham and Pooley, Acorn and Gupta, Agrim and Mandlekar, Ajay and Jain, Ajinkya and others},
  booktitle=ICRA,
  year={2024},
}

@inproceedings{octo,
    title={Octo: An Open-Source Generalist Robot Policy},
    author = {Ghosh, Dibya and Walke, Homer Rich and Pertsch, Karl and Black, Kevin and Mees, Oier and Dasari, Sudeep and Hejna, Joey and Kreiman, Tobias and Xu, Charles and Luo, Jianlan and Tan, You Liang and Chen, Lawrence Yunliang and Vuong, Quan and Xiao, Ted and Sanketi, Pannag R and Sadigh, Dorsa and Finn, Chelsea and Levine, Sergey},
    booktitle = RSS,
    year = {2024}
}

@InProceedings{openvla,
  title = 	 {OpenVLA: An Open-Source Vision-Language-Action Model},
  author =       {Kim, Moo Jin and Pertsch, Karl and Karamcheti, Siddharth and Xiao, Ted and Balakrishna, Ashwin and Nair, Suraj and Rafailov, Rafael and Foster, Ethan P and Sanketi, Pannag R and Vuong, Quan and Kollar, Thomas and Burchfiel, Benjamin and Tedrake, Russ and Sadigh, Dorsa and Levine, Sergey and Liang, Percy and Finn, Chelsea},
  booktitle = 	 CORL,
  year = 	 {2025},
}

@misc{pizero,
      title={$\pi_0$: A Vision-Language-Action Flow Model for General Robot Control}, 
      author={Kevin Black and Noah Brown and Danny Driess and Adnan Esmail and Michael Equi and Chelsea Finn and Niccolo Fusai and Lachy Groom and Karol Hausman and Brian Ichter and Szymon Jakubczak and Tim Jones and Liyiming Ke and Sergey Levine and Adrian Li-Bell and Mohith Mothukuri and Suraj Nair and Karl Pertsch and Lucy Xiaoyang Shi and James Tanner and Quan Vuong and Anna Walling and Haohuan Wang and Ury Zhilinsky},
      year={2024},
      howpublished={\url{https://arxiv.org/abs/2410.24164}}
}

@INPROCEEDINGS{rt1, 
    author    = {Anthony Brohan AND Noah Brown AND Justice Carbajal AND Yevgen Chebotar AND Joseph Dabis AND Chelsea Finn AND Keerthana Gopalakrishnan AND Karol Hausman AND Alexander Herzog AND Jasmine Hsu AND Julian Ibarz AND Brian Ichter AND Alex Irpan AND Tomas Jackson AND Sally Jesmonth AND Nikhil Joshi AND Ryan Julian AND Dmitry Kalashnikov AND Yuheng Kuang AND Isabel Leal AND Kuang-Huei Lee AND Sergey Levine AND Yao Lu AND Utsav Malla AND Deeksha Manjunath AND Igor Mordatch AND Ofir Nachum AND Carolina Parada AND Jodilyn  Peralta AND Emily Perez AND Karl Pertsch AND Jornell  Quiambao AND Kanishka Rao AND Michael S Ryoo AND Grecia  Salazar AND Pannag R Sanketi AND Kevin  Sayed AND Jaspiar  Singh AND Sumedh  Sontakke AND Austin  Stone AND Clayton  Tan AND Huong  Tran AND Vincent Vanhoucke AND Steve  Vega AND Quan H Vuong AND Fei Xia AND Ted Xiao AND Peng Xu AND Sichun Xu AND Tianhe Yu AND Brianna  Zitkovich}, 
    title     = {RT-1: Robotics Transformer for Real-World Control at Scale}, 
    booktitle = RSS, 
    year      = {2023}
}

@InProceedings{rt2,
  title = 	 {RT-2: Vision-Language-Action Models Transfer Web Knowledge to Robotic Control},
  author =       {Zitkovich, Brianna and Yu, Tianhe and Xu, Sichun and Xu, Peng and Xiao, Ted and Xia, Fei and Wu, Jialin and Wohlhart, Paul and Welker, Stefan and Wahid, Ayzaan and Vuong, Quan and Vanhoucke, Vincent and Tran, Huong and Soricut, Radu and Singh, Anikait and Singh, Jaspiar and Sermanet, Pierre and Sanketi, Pannag R. and Salazar, Grecia and Ryoo, Michael S. and Reymann, Krista and Rao, Kanishka and Pertsch, Karl and Mordatch, Igor and Michalewski, Henryk and Lu, Yao and Levine, Sergey and Lee, Lisa and Lee, Tsang-Wei Edward and Leal, Isabel and Kuang, Yuheng and Kalashnikov, Dmitry and Julian, Ryan and Joshi, Nikhil J. and Irpan, Alex and Ichter, Brian and Hsu, Jasmine and Herzog, Alexander and Hausman, Karol and Gopalakrishnan, Keerthana and Fu, Chuyuan and Florence, Pete and Finn, Chelsea and Dubey, Kumar Avinava and Driess, Danny and Ding, Tianli and Choromanski, Krzysztof Marcin and Chen, Xi and Chebotar, Yevgen and Carbajal, Justice and Brown, Noah and Brohan, Anthony and Arenas, Montserrat Gonzalez and Han, Kehang},
  booktitle = 	 CORL,
  year = 	 {2023}
}

@article{vr-goggles,
  title={Vr-goggles for robots: Real-to-sim domain adaptation for visual control},
  author={Zhang, Jingwei and Tai, Lei and Yun, Peng and Xiong, Yufeng and Liu, Ming and Boedecker, Joschka and Burgard, Wolfram},
  journal=ieeeral,
  volume={4},
  number={2},
  year={2019},
  publisher={IEEE}
}

@inproceedings{viola,
  title={Viola: Imitation learning for vision-based manipulation with object proposal priors},
  author={Zhu, Yifeng and Joshi, Abhishek and Stone, Peter and Zhu, Yuke},
  booktitle=CORL,
  year={2023},
}

@InProceedings{dalal2023imitating,
  title = 	 {Imitating Task and Motion Planning with Visuomotor Transformers},
  author =       {Dalal, Murtaza and Mandlekar, Ajay and Garrett, Caelan Reed and Handa, Ankur and Salakhutdinov, Ruslan and Fox, Dieter},
  booktitle = 	 CORL,
  year = 	 {2023}
}

@inproceedings{sam2,
title={{SAM} 2: Segment Anything in Images and Videos},
author={Nikhila Ravi and Valentin Gabeur and Yuan-Ting Hu and Ronghang Hu and Chaitanya Ryali and Tengyu Ma and Haitham Khedr and Roman R{\"a}dle and Chloe Rolland and Laura Gustafson and Eric Mintun and Junting Pan and Kalyan Vasudev Alwala and Nicolas Carion and Chao-Yuan Wu and Ross Girshick and Piotr Dollar and Christoph Feichtenhofer},
booktitle=ICLR,
year={2025}
}

@INPROCEEDINGS{droiddataset, 
    AUTHOR    = {Alexander Khazatsky AND Karl Pertsch AND Suraj Nair AND Ashwin Balakrishna AND Sudeep Dasari AND Siddharth Karamcheti AND Soroush Nasiriany AND Mohan Kumar Srirama AND Lawrence Yunliang Chen AND Kirsty Ellis AND Peter David Fagan AND Joey Hejna AND Masha Itkina AND Marion Lepert AND Yecheng Jason Ma AND Patrick Tree Miller AND Jimmy Wu AND Suneel Belkhale AND Shivin Dass AND Huy Ha AND Arhan Jain AND Abraham Lee AND Youngwoon Lee AND Marius Memmel AND Sungjae Park AND Ilija Radosavovic AND Kaiyuan Wang AND Albert Zhan AND Kevin Black AND Cheng Chi AND Kyle Beltran Hatch AND Shan Lin AND Jingpei Lu AND Jean Mercat AND Abdul Rehman AND Pannag R Sanketi AND Archit Sharma AND Cody Simpson AND Quan Vuong AND Homer Rich Walke AND Blake Wulfe AND Ted Xiao AND Jonathan Heewon Yang AND Arefeh Yavary AND Tony Z. Zhao AND Christopher Agia AND Rohan Baijal AND Mateo Guaman Castro AND Daphne Chen AND Qiuyu Chen AND Trinity Chung AND Jaimyn Drake AND Ethan Paul Foster AND Jensen Gao AND David Antonio Herrera AND Minho Heo AND Kyle Hsu AND Jiaheng Hu AND Donovon Jackson AND Charlotte Le AND Yunshuang Li AND Roy Lin AND Zehan Ma AND Abhiram Maddukuri AND Suvir Mirchandani AND Daniel Morton AND Tony Nguyen AND Abigail O'Neill AND Rosario Scalise AND Derick Seale AND Victor Son AND Stephen Tian AND Emi Tran AND Andrew E. Wang AND Yilin Wu AND Annie Xie AND Jingyun Yang AND Patrick Yin AND Yunchu Zhang AND Osbert Bastani AND Glen Berseth AND Jeannette Bohg AND Ken Goldberg AND Abhinav Gupta AND Abhishek Gupta AND Dinesh Jayaraman AND Joseph J Lim AND Jitendra Malik AND Roberto Martín-Martín AND Subramanian Ramamoorthy AND Dorsa Sadigh AND Shuran Song AND Jiajun Wu AND Michael C. Yip AND Yuke Zhu AND Thomas Kollar AND Sergey Levine AND Chelsea Finn}, 
    TITLE     = {{DROID}: A Large-Scale In-The-Wild Robot Manipulation Dataset}, 
    BOOKTITLE = RSS, 
    YEAR      = {2024}
}

@INPROCEEDINGS{decomposegengap,
  author={Xie, Annie and Lee, Lisa and Xiao, Ted and Finn, Chelsea},
  booktitle=ICRA, 
  title={Decomposing the Generalization Gap in Imitation Learning for Visual Robotic Manipulation}, 
  year={2024}
}

@InProceedings{groundingdino,
    author="Liu, Shilong
    and Zeng, Zhaoyang
    and Ren, Tianhe
    and Li, Feng
    and Zhang, Hao
    and Yang, Jie
    and Jiang, Qing
    and Li, Chunyuan
    and Yang, Jianwei
    and Su, Hang
    and Zhu, Jun
    and Zhang, Lei",
    title="Grounding {DINO}: Marrying {DINO} with Grounded Pre-training for Open-Set Object Detection",
    booktitle=ECCV,
    year={2025}
}

@inproceedings{Dosovitskiy.2021,
title={An Image is Worth 16x16 Words: Transformers for Image Recognition at Scale},
author={Alexey Dosovitskiy and Lucas Beyer and Alexander Kolesnikov and Dirk Weissenborn and Xiaohua Zhai and Thomas Unterthiner and Mostafa Dehghani and Matthias Minderer and Georg Heigold and Sylvain Gelly and Jakob Uszkoreit and Neil Houlsby},
booktitle=ICLR,
year={2021}
}

@inproceedings{Ho.2020,
  title = {Denoising Diffusion Probabilistic Models},
  booktitle = neurips,
  author = {Ho, Jonathan and Jain, Ajay and Abbeel, Pieter},
  year = {2020}
}

@inproceedings{Vaswani.2017,
  title={Attention Is All You Need},
  author={Vaswani, Ashish and Shazeer, Noam and Parmar, Niki and Uszkoreit, Jakob and Jones, Llion and Gomez, Aidan N. and Kaiser, {\L}ukasz and Polosukhin, Illia},
  booktitle = neurips,
  year={2017}
}

@InProceedings{Walke.2023,
  title = 	 {BridgeData V2: A Dataset for Robot Learning at Scale},
  author =       {Walke, Homer Rich and Black, Kevin and Zhao, Tony Z. and Vuong, Quan and Zheng, Chongyi and Hansen-Estruch, Philippe and He, Andre Wang and Myers, Vivek and Kim, Moo Jin and Du, Max and Lee, Abraham and Fang, Kuan and Finn, Chelsea and Levine, Sergey},
  booktitle = 	 CORL,
  year = 	 {2023},
}

@INPROCEEDINGS{Fang.2024,
  author={Fang, Hao-Shu and Fang, Hongjie and Tang, Zhenyu and Liu, Jirong and Wang, Chenxi and Wang, Junbo and Zhu, Haoyi and Lu, Cewu},
  booktitle= ICRA, 
  title={RH20T: A Comprehensive Robotic Dataset for Learning Diverse Skills in One-Shot}, 
  year={2024}
}

@INPROCEEDINGS{roboagent,
  author={Bharadhwaj, Homanga and Vakil, Jay and Sharma, Mohit and Gupta, Abhinav and Tulsiani, Shubham and Kumar, Vikash},
  booktitle= ICRA, 
  title={{RoboAgent}: Generalization and Efficiency in Robot Manipulation via Semantic Augmentations and Action Chunking}, 
  year={2024}
}

@INPROCEEDINGS{robomind,
  author = {Wu, Kun and Hou, Chengkai and Liu, Jiaming and Che, Zhengping and Ju, Xiaozhu and Yang, Zhuqin and Li, Meng and Zhao, Yinuo and Xu, Zhiyuan and Yang, Guang and Fan, Shichao and Wang, Xinhua and Liao, Fei and Zhao, Zhen and Li, Guangyu and Jin, Zhao and Wang, Lecheng and Mao, Jilei and Liu, Ning and Ren, Pei and Zhang, Qiang and Lyu, Yaoxu and Liu, Mengzhen and Jingyang, He and Luo, Yulin and Gao, Zeyu and Li, Chenxuan and Gu, Chenyang and Fu, Yankai and Wu, Di and Wang, Xingyu and Chen, Sixiang and Wang, Zhenyu and An, Pengju and Qian, Siyuan and Zhang, Shanghang and Tang, Jian},
  booktitle= RSS, 
  title={{RoboMIND}: Benchmark on Multi-embodiment Intelligence Normative Data for Robot Manipulation}, 
  year={2025}
}

@InProceedings{Fu.2024,
  title = 	 {Mobile {ALOHA}: Learning Bimanual Mobile Manipulation using Low-Cost Whole-Body Teleoperation},
  author =       {Fu, Zipeng and Zhao, Tony Z. and Finn, Chelsea},
  booktitle = 	 CORL,
  year = 	 {2025}
}

@InProceedings{Li.2024,
  title = 	 {Evaluating Real-World Robot Manipulation Policies in Simulation},
  author =       {Li, Xuanlin and Hsu, Kyle and Gu, Jiayuan and Mees, Oier and Pertsch, Karl and Walke, Homer Rich and Fu, Chuyuan and Lunawat, Ishikaa and Sieh, Isabel and Kirmani, Sean and Levine, Sergey and Wu, Jiajun and Finn, Chelsea and Su, Hao and Vuong, Quan and Xiao, Ted},
  booktitle = 	 CORL,
  year = 	 {2025}
}

@misc{gpt-4o,
  title={GPT-4 Technical Report}, 
  author={OpenAI and Josh Achiam and Steven Adler and Sandhini Agarwal and Lama Ahmad and Ilge Akkaya and Florencia Leoni Aleman and Diogo Almeida and Janko Altenschmidt and Sam Altman and Shyamal Anadkat and Red Avila and Igor Babuschkin and Suchir Balaji and Valerie Balcom and Paul Baltescu and Haiming Bao and Mohammad Bavarian and Jeff Belgum and Irwan Bello and Jake Berdine and Gabriel Bernadett-Shapiro and Christopher Berner and Lenny Bogdonoff and Oleg Boiko and Madelaine Boyd and Anna-Luisa Brakman and Greg Brockman and Tim Brooks and Miles Brundage and Kevin Button and Trevor Cai and Rosie Campbell and Andrew Cann and Brittany Carey and Chelsea Carlson and Rory Carmichael and Brooke Chan and Che Chang and Fotis Chantzis and Derek Chen and Sully Chen and Ruby Chen and Jason Chen and Mark Chen and Ben Chess and Chester Cho and Casey Chu and Hyung Won Chung and Dave Cummings and Jeremiah Currier and Yunxing Dai and Cory Decareaux and Thomas Degry and Noah Deutsch and Damien Deville and Arka Dhar and David Dohan and Steve Dowling and Sheila Dunning and Adrien Ecoffet and Atty Eleti and Tyna Eloundou and David Farhi and Liam Fedus and Niko Felix and Simón Posada Fishman and Juston Forte and Isabella Fulford and Leo Gao and Elie Georges and Christian Gibson and Vik Goel and Tarun Gogineni and Gabriel Goh and Rapha Gontijo-Lopes and Jonathan Gordon and Morgan Grafstein and Scott Gray and Ryan Greene and Joshua Gross and Shixiang Shane Gu and Yufei Guo and Chris Hallacy and Jesse Han and Jeff Harris and Yuchen He and Mike Heaton and Johannes Heidecke and Chris Hesse and Alan Hickey and Wade Hickey and Peter Hoeschele and Brandon Houghton and Kenny Hsu and Shengli Hu and Xin Hu and Joost Huizinga and Shantanu Jain and Shawn Jain and Joanne Jang and Angela Jiang and Roger Jiang and Haozhun Jin and Denny Jin and Shino Jomoto and Billie Jonn and Heewoo Jun and Tomer Kaftan and Łukasz Kaiser and Ali Kamali and Ingmar Kanitscheider and Nitish Shirish Keskar and Tabarak Khan and Logan Kilpatrick and Jong Wook Kim and Christina Kim and Yongjik Kim and Jan Hendrik Kirchner and Jamie Kiros and Matt Knight and Daniel Kokotajlo and Łukasz Kondraciuk and Andrew Kondrich and Aris Konstantinidis and Kyle Kosic and Gretchen Krueger and Vishal Kuo and Michael Lampe and Ikai Lan and Teddy Lee and Jan Leike and Jade Leung and Daniel Levy and Chak Ming Li and Rachel Lim and Molly Lin and Stephanie Lin and Mateusz Litwin and Theresa Lopez and Ryan Lowe and Patricia Lue and Anna Makanju and Kim Malfacini and Sam Manning and Todor Markov and Yaniv Markovski and Bianca Martin and Katie Mayer and Andrew Mayne and Bob McGrew and Scott Mayer McKinney and Christine McLeavey and Paul McMillan and Jake McNeil and David Medina and Aalok Mehta and Jacob Menick and Luke Metz and Andrey Mishchenko and Pamela Mishkin and Vinnie Monaco and Evan Morikawa and Daniel Mossing and Tong Mu and Mira Murati and Oleg Murk and David Mély and Ashvin Nair and Reiichiro Nakano and Rajeev Nayak and Arvind Neelakantan and Richard Ngo and Hyeonwoo Noh and Long Ouyang and Cullen O'Keefe and Jakub Pachocki and Alex Paino and Joe Palermo and Ashley Pantuliano and Giambattista Parascandolo and Joel Parish and Emy Parparita and Alex Passos and Mikhail Pavlov and Andrew Peng and Adam Perelman and Filipe de Avila Belbute Peres and Michael Petrov and Henrique Ponde de Oliveira Pinto and Michael and Pokorny and Michelle Pokrass and Vitchyr H. Pong and Tolly Powell and Alethea Power and Boris Power and Elizabeth Proehl and Raul Puri and Alec Radford and Jack Rae and Aditya Ramesh and Cameron Raymond and Francis Real and Kendra Rimbach and Carl Ross and Bob Rotsted and Henri Roussez and Nick Ryder and Mario Saltarelli and Ted Sanders and Shibani Santurkar and Girish Sastry and Heather Schmidt and David Schnurr and John Schulman and Daniel Selsam and Kyla Sheppard and Toki Sherbakov and Jessica Shieh and Sarah Shoker and Pranav Shyam and Szymon Sidor and Eric Sigler and Maddie Simens and Jordan Sitkin and Katarina Slama and Ian Sohl and Benjamin Sokolowsky and Yang Song and Natalie Staudacher and Felipe Petroski Such and Natalie Summers and Ilya Sutskever and Jie Tang and Nikolas Tezak and Madeleine B. Thompson and Phil Tillet and Amin Tootoonchian and Elizabeth Tseng and Preston Tuggle and Nick Turley and Jerry Tworek and Juan Felipe Cerón Uribe and Andrea Vallone and Arun Vijayvergiya and Chelsea Voss and Carroll Wainwright and Justin Jay Wang and Alvin Wang and Ben Wang and Jonathan Ward and Jason Wei and CJ Weinmann and Akila Welihinda and Peter Welinder and Jiayi Weng and Lilian Weng and Matt Wiethoff and Dave Willner and Clemens Winter and Samuel Wolrich and Hannah Wong and Lauren Workman and Sherwin Wu and Jeff Wu and Michael Wu and Kai Xiao and Tao Xu and Sarah Yoo and Kevin Yu and Qiming Yuan and Wojciech Zaremba and Rowan Zellers and Chong Zhang and Marvin Zhang and Shengjia Zhao and Tianhao Zheng and Juntang Zhuang and William Zhuk and Barret Zoph},
  howpublished={\url{https://arxiv.org/abs/2303.08774}},
  year={2024}
}

@INPROCEEDINGS{mujoco,
  author={Todorov, Emanuel and Erez, Tom and Tassa, Yuval},
  booktitle= IROS, 
  title={MuJoCo: A physics engine for model-based control}, 
  year={2012}
}

@inproceedings{Devin.2018,
  title = {Deep Object-Centric Representations for Generalizable Robot Learning},
  booktitle = ICRA,
  author = {Devin, Coline and Abbeel, Pieter and Darrell, Trevor and Levine, Sergey},
  year = {2018}
}

@inproceedings{Sieb.2020,
  title = {Graph-Structured Visual Imitation},
  booktitle = CORL,
  author = {Sieb, Maximilian and Xian, Zhou and Huang, Audrey and Kroemer, Oliver and Fragkiadaki, Katerina},
  year = {2020}
}

@inproceedings{Zeng.2021,
  title = {Transporter Networks: Rearranging the Visual World for Robotic Manipulation},
  booktitle = CORL,
  author = {Zeng, Andy and Florence, Pete and Tompson, Jonathan and Welker, Stefan and Chien, Jonathan and Attarian, Maria and Armstrong, Travis and Krasin, Ivan and Duong, Dan and Sindhwani, Vikas and Lee, Johnny},
  year = {2021}
}

@inproceedings{Nair.2022,
  title = {{R3M}: A Universal Visual Representation for Robot Manipulation},
  booktitle = CORL,
  author = {Nair, Suraj and Rajeswaran, Aravind and Kumar, Vikash and Finn, Chelsea and Gupta, Abhinav},
  year = {2022}
}

@inproceedings{Shridhar.2022,
  title = {{CLIPort}: What and Where Pathways for Robotic Manipulation},
  booktitle = CORL,
  author = {Shridhar, Mohit and Manuelli, Lucas and Fox, Dieter},
  year = {2022}
}

@inproceedings{Chen.2024,
  title = {{{SUGAR}}: Pre-training 3D Visual Representations for Robotics},
  booktitle = ICCV,
  author = {Chen, Shizhe and Garcia, Ricardo and Laptev, Ivan and Schmid, Cordelia},
  year = {2024}
}

@inproceedings{maniskill,
  title={{ManiSkill2}: A Unified Benchmark for Generalizable Manipulation Skills},
  author={Jiayuan Gu and Fanbo Xiang and Xuanlin Li and Zhan Ling and Xiqiang Liu and Tongzhou Mu and Yihe Tang and Stone Tao and Xinyue Wei and Yunchao Yao and Xiaodi Yuan and Pengwei Xie and Zhiao Huang and Rui Chen and Hao Su},
  booktitle=ICLR,
  year={2023}
}

@inproceedings{Karamcheti.2023,
  title={Language-Driven Representation Learning for Robotics},
  author={Siddharth Karamcheti and Suraj Nair and Annie S. Chen and Thomas Kollar and Chelsea Finn and Dorsa Sadigh and Percy Liang},
  booktitle=RSS,
  year={2023}
}

@inproceedings{Manuelli.2022,
  title = {{KPAM}: KeyPoint Affordances for Category-Level Robotic Manipulation},
  booktitle = ISRR,
  author = {Manuelli, Lucas and Gao, Wei and Florence, Peter and Tedrake, Russ},
  year = {2019}
}

@inproceedings{Huang.2024,
  title = {{ReKep}: Spatio-Temporal Reasoning of Relational Keypoint Constraints for Robotic Manipulation},
  booktitle = CORL,
  author = {Huang, Wenlong and Wang, Chen and Li, Yunzhu and Zhang, Ruohan and {Fei-Fei}, Li},
  year = {2024}
}

@inproceedings{Nasiriany.2024,
  title = {{PIVOT}: Iterative Visual Prompting Elicits Actionable Knowledge for {{VLMs}}},
  booktitle = ICML,
  author = {Nasiriany, Soroush and Xia, Fei and Yu, Wenhao and Xiao, Ted and Liang, Jacky and Dasgupta, Ishita and Xie, Annie and Driess, Danny and Wahid, Ayzaan and Xu, Zhuo and Vuong, Quan and Zhang, Tingnan and Lee, Tsang-Wei Edward and Lee, Kuang-Huei and Xu, Peng and Kirmani, Sean and Zhu, Yuke and Zeng, Andy and Hausman, Karol and Heess, Nicolas and Finn, Chelsea and Levine, Sergey and Ichter, Brian},
  year = {2024}
}

@article{fangandliu2024moka,
      title={{MOKA}: Open-World Robotic Manipulation through Mark-Based Visual Prompting},
      author={Kuan Fang and Fangchen Liu and Pieter Abbeel and Sergey Levine},
      journal=RSS,
      year={2024}
  }

@inproceedings{Tobin.2017,
  title = {Domain Randomization for Transferring Deep Neural Networks from Simulation to the Real World},
  booktitle = IROS,
  author = {Tobin, Josh and Fong, Rachel and Ray, Alex and Schneider, Jonas and Zaremba, Wojciech and Abbeel, Pieter},
  year = {2017}
}

@InProceedings{saga,
  title = 	 {Enhancing Visual Domain Robustness in Behaviour Cloning via Saliency-Guided Augmentation},
  author =       {Zhuang, Zheyu and WANG, RUIYU and Ingelhag, Nils and Kyrki, Ville and Kragic, Danica},
  booktitle = 	 CORL,
  year = 	 {2025}
}

@misc{rcs,
  title={{Robot Control Stack}: {A} Lean Ecosystem for Robot Learning at Scale}, 
  author={Tobias J{\"u}lg and Pierre Krack and Seongjin Bien and Yannik Blei and Khaled Gamal and Ken Nakahara and Johannes Hechtl and Roberto Calandra and Wolfram Burgard and Florian Walter},
  year={2025},
  howpublished = {\url{https://arxiv.org/abs/2509.14932}}
}

\end{document}